\documentclass{article}

\usepackage{microtype}
\usepackage{graphicx}
\usepackage{booktabs} 
\usepackage{bbm} 
\usepackage{dsfont}  
\usepackage{caption}
\usepackage{subcaption}     

\usepackage{amsmath,amsfonts,amsbsy,amssymb,amsthm,bm,calc,lipsum,array,tabularx}
\usepackage{multirow,multicol,changepage,pifont,enumitem,mathtools}
\usepackage{algorithm, algcompatible}
\usepackage[noend]{algpseudocode}

\makeatletter
\@namedef{ver@algorithmic.sty}{}
\makeatother

\usepackage{wrapfig,caption}
\usepackage[dvipsnames]{xcolor} 
\usepackage[mathscr]{euscript}
\usepackage{makecell}

\newcommand{\norm}[1]{\Vert {#1} \Vert}

\newcommand{\R}{\mathbb{R}}

\newcommand{\E}{\mathbb{E}}

\DeclareMathOperator*{\argmax}{arg\,max}
\DeclareMathOperator*{\argmin}{arg\,min}

\DeclarePairedDelimiterX{\infdivx}[2]{(}{)}{%
  #1\;\delimsize\|\;#2%
}
\newcommand{\DKL}{\mathrm{D}_{\rm KL}\infdivx}

\newcommand{\given}{\,|\,}

\newcommand{\supp}{\mathrm{supp}}

\newcommand{\env}[1]{\texttt{#1}}
\newcommand{\strat}[1]{\texttt{#1}}
\newcommand{\success}[1]{S_{#1}}

\newcommand{\ag}{\textsc{ag}}

\definecolor{coolestcolor}{RGB}{222,44,111}
\definecolor{tbdblue}{RGB}{70,70,200}

\newtheorem{proposition}{Proposition}

\def\-{\,\text{--}\,}

\newcolumntype{Y}{>{\centering\arraybackslash}X} 
\newcolumntype{L}[1]{>{\raggedright\let\newline\\\arraybackslash\hspace{0pt}}m{#1}} 
\newcolumntype{C}[1]{>{\centering\arraybackslash\hspace{0pt}}m{#1}} 

\usepackage{hyperref}

\usepackage[accepted]{icml2020_style/icml2020}

\icmltitlerunning{Maximum Entropy Gain Exploration for Long Horizon Multi-goal Reinforcement Learning}

\usepackage{subfiles} 

\begin{document}

\twocolumn[
\icmltitle{
	Maximum Entropy Gain Exploration for \\Long Horizon Multi-goal Reinforcement Learning
	}


\icmlsetsymbol{equal}{*}

\begin{icmlauthorlist}
\icmlauthor{Silviu Pitis}{equal,to,vec}
\icmlauthor{Harris Chan}{equal,to,vec}
\icmlauthor{Stephen Zhao}{to}
\icmlauthor{Bradly Stadie}{vec}
\icmlauthor{Jimmy Ba}{to,vec}
\end{icmlauthorlist}

\icmlaffiliation{to}{University of Toronto}
\icmlaffiliation{vec}{Vector Institute}

\icmlcorrespondingauthor{Silviu Pitis}{spitis@cs.toronto.edu}
\icmlcorrespondingauthor{Harris Chan}{hchan@cs.toronto.edu}

\icmlkeywords{Machine Learning, ICML}

\vskip 0.3in
]



\printAffiliationsAndNotice{\icmlEqualContribution} 

\begin{abstract}
What goals should a multi-goal reinforcement learning agent pursue during training in long-horizon tasks? When the desired (test time) goal distribution is too distant to offer a useful learning signal, we argue that the agent should not pursue unobtainable goals. Instead, it should set its own intrinsic goals that maximize the entropy of the historical achieved goal distribution. We propose to optimize this objective by having the agent pursue past achieved goals in sparsely explored areas of the goal space, which focuses exploration on the frontier of the achievable goal set.
We show that our strategy achieves an order of magnitude better sample efficiency than the prior state of the art on long-horizon multi-goal tasks including maze navigation and block stacking.\footnotemark
\end{abstract}

\section{Introduction}\label{section_introduction}

Multi-goal reinforcement learning (RL) agents \cite{plappert2018multi,schaul2015prioritized,kaelbling1993learning} learn goal-conditioned behaviors that can achieve and generalize across a range of different goals. 
Multi-goal RL forms a core component of hierarchical agents \cite{sutton1999between,nachum2018data}, and has been shown to allow unsupervised agents to learn useful skills for downstream tasks \cite{warde-farley2018unsupervised,hansen2019fast}. Recent advances in goal relabeling \cite{andrychowicz2017hindsight} have made learning possible in complex, sparse-reward environments whose goal spaces are either dense in the initial state distribution \cite{plappert2018multi} or structured as a curriculum \cite{colas2018curious}. But learning without demonstrations in 
long-horizon tasks remains a challenge \cite{nair2018overcoming,trott2019keeping}, as learning signal decreases exponentially with the horizon \cite{osband2014generalization}. 

In this paper, we improve upon existing approaches to intrinsic goal setting and show how multi-goal agents can form an automatic behavioural goal curriculum that allows them to master long-horizon, sparse reward tasks.
We begin with an algorithmic framework for goal-seeking agents that contextualizes  prior work \cite{baranes2013active,florensa2018automatic,warde-farley2018unsupervised,nair2018visual,pong2019skew} and argue that past goal selection mechanisms are not well suited for long-horizon, sparse reward tasks  (Section \ref{section_background}). By framing the long-horizon goal seeking task as optimizing an initially ill-conditioned distribution matching objective \cite{lee2019efficient}, we arrive at our unsupervised Maximum Entropy Goal Achievement (MEGA) objective, which maximizes the entropy of the past achieved goal set. This early unsupervised objective is annealed into the original supervised objective once the latter becomes tractable, resulting in our OMEGA objective (Section \ref{section_method}).\footnotetext{Code available at \url{https://github.com/spitis/mrl}}

We propose a practical algorithmic approach to maximizing entropy, which pursues past achieved goals in sparsely explored areas of the achieved goal distribution, as measured by a learned density model.
The agent revisits and explores around these areas, pushing the frontier of achieved goals forward \cite{ecoffet2019go}. 
This strategy, similar in spirit to \citet{baranes2013active} and \citet{florensa2018automatic}, encourages the agent to explore at the edge of its abilities, which avoids spending environment steps in pursuit of already mastered or unobtainable goals. 
When used in combination with hindsight experience replay and an off-policy learning algorithm, our method achieves more than an order of magnitude better sample efficiency than the prior state of the art on difficult exploration tasks, including long-horizon mazes and block stacking (Section \ref{section_empirical}). Finally, we draw connections between our approach and the empowerment objective \cite{klyubin2005empowerment,salge2014empowerment} and identify a key difference to prior work: rather than maximize empowerment on-policy by setting maximally diverse goals during training \cite{gregor2016variational,warde-farley2018unsupervised,nair2018visual,pong2019skew}, our proposed approach maximizes it off-policy by setting goals on the frontier of the past achieved goal set. We conclude with discussion of related and future work (Sections \ref{section_related}-\ref{section_conclusion}).
\vfill

\begin{figure*}[!t]
\centering
\includegraphics[width=6.4in]{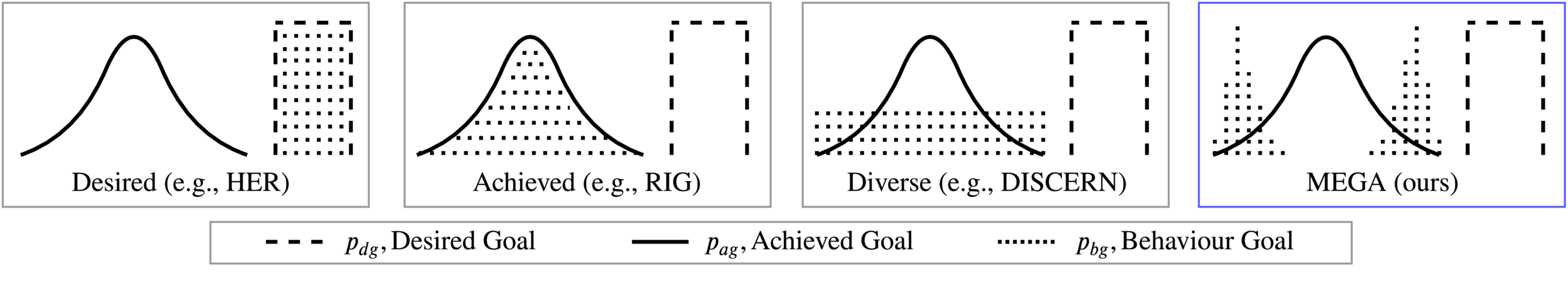}
\vspace{-0.16in}
\caption{Illustration of density-based \textsc{Select} mechanisms at start of training, when achieved ($p_{ag}$) and desired ($p_{dg}$) goal distributions are disconnected. HER samples goals from the desired distribution $p_{dg}$. RIG samples from the achieved distribution $p_{ag}$. DISCERN and Skew-Fit skew $p_{ag}$ to sample diverse achieved goals. Our approach (MEGA) focuses on low density regions of $p_{ag}$. See Subsection \ref{subsection_setting_intrinsic_goals}.} \label{fig_density_based_select_comparison}
\end{figure*}

\section{The Long-Horizon Problem}\label{section_background}

\subsection{Preliminaries}

We consider the multi-goal reinforcement learning (RL) setting, described by a generalized Markov Decision Process (MDP) $\mathcal{M} = \langle S, A, T, G, [p_{dg}] \rangle$, where $S$, $A$, $T$, and $G$ are the state space, action space, transition function and goal space, respectively \cite{sutton2018reinforcement,schaul2015universal} and $p_{dg}$ is an optional desired goal distribution. In the most general version of this problem each goal is a tuple $g = \langle R_g, \gamma_g \rangle$, where $R_g: S \to \mathbb{R}$ is a reward function and $\gamma_g \in [0, 1]$ is a discount factor \cite{sutton2011horde}, so that ``solving'' goal $g \in G$ amounts to finding an optimal policy in the classical MDP $\mathcal{M}_g = \langle S, A, T, R_g, \gamma_g \rangle$. Although goal-oriented methods are general and could be applied to dense reward MDPs (including the standard RL problem, as done by  \citet{warde-farley2018unsupervised}, among others), we restrict our present attention to the sparse reward case, where each goal $g$ corresponds to a set of ``success'' states, $\success{g}$, with $R_g: S \to \{-1, 0\}$ \cite{plappert2018multi} defined as $R_g(s) = \mathbb{I}\{s \in \success{g}\} + c$.  Following \citeauthor{plappert2018multi}, we use base reward $c = -1$, which typically leads to more stable training than the more natural $c = 0$ (see \citet{van2019using} for a possible explanation). We also adopt the usual form $\success{g} = \{s \given d(\ag(s), g) < \epsilon \}$, where $\ag: S \to G$ maps state $s$ to an ``achieved goal'' $\ag(s)$ and $d$ is a metric on $G$. An agent's ``achieved goal distribution'' $p_{ag}$ is the distribution of goals achieved by states $s$ (i.e., $\ag(s)$) the agent visits (not necessarily the final state in a trajectory). Note that this may be on-policy (induced by the current policy) or historical, as we will specify below. The agent must learn to achieve success and, if the environment is not episodic, maintain it. In the episodic case, we can think of each goal $g \in G$ as specifying a skill or option $o \in \Omega$ \cite{sutton1999between,eysenbach2018diversity}, so that multi-goal reinforcement learning is closely related to hierarchical reinforcement learning \cite{nachum2018data}.

A common approach to multi-goal RL, which we adopt, trains a goal-conditioned state-action value function, $Q: S \times A \times G \to \R$, using an off-policy learning algorithm that can leverage data from other policies (past and exploratory) to optimize the current policy \cite{schaul2015prioritized}. A goal-conditioned policy, $\pi: S \times G \to A$, is either induced via greedy action selection \cite{mnih2013playing} or learned using policy gradients. Noise is added to $\pi$ during exploration to form exploratory policy $\pi_\textrm{explore}$. Our continuous control experiments all use the DDPG algorithm \cite{lillicrap2015continuous}, which parameterizes actor and critic separately, and trains both concurrently using Q-learning for the critic \cite{watkins1992q}, and deterministic policy gradients \cite{silver2014deterministic} for the actor. DDPG uses a replay buffer to store past experiences, which is then sampled from to train the actor and critic networks.

\subsection{Sparse rewards and the long horizon problem}

\newcommand{\Var}[1]{\texttt{#1}}
\begin{algorithm*}
	\caption{Unified Framework for Multi-goal Agents} \label{alg_goal_seeking}
	\begin{small}\vspace{-0.15in}
	\begin{multicols}{2}\noindent
		\begin{algorithmic}
		\Function{Train}{$*args$}:
			\begin{adjustwidth}{\algorithmicindent}{0pt}
		    Alternate between collecting experience using \textsc{Rollout} and optimizing the parameters using \textsc{Optimize}.
		    \end{adjustwidth}
		\EndFunction
		\Statex
		\Function{Rollout }{policy $\pi_\textrm{explore}$, buffer $\mathcal{B}$, $*args$}:
    		\State $\Var{g} \gets \textsc{Select}( *args)$
    		\State $\Var{s}_0 \gets $ initial state
    		\For{$t$ in $0 \dots T-1$}
    			\State $\Var{a}_t, \Var{s}_{t+1} \gets $ execute $\pi_\textrm{explore}(\Var{s}_t, \Var{g})$ in environment
    			\State $\Var{r}_t \gets \textsc{Reward}(\Var{s}_t, \Var{a}_t, \Var{s}_{t+1}, \Var{g})$
    			\State Store $(\Var{s}_t, \Var{a}_t, \Var{s}_{t+1}, \Var{r}_t, \Var{g})$ in replay buffer $\mathcal{B}$
    		\EndFor
    	\EndFunction
		\Statex
		\Function{Optimize }{buffer $\mathcal{B}$, algorithm $\mathcal{A}$, parameters $\theta$}:
		    \State Sample mini-batch $B = \{(\Var{s}, \Var{a}, \Var{s'}, \Var{r}, \Var{g})_i\}_{i=1}^{N} \sim \mathcal{B}$
		    \State $B' \gets \textsc{Relabel}(B, *args)$
		    \State Optimize $\theta$ using $\mathcal{A}$ (e.g., DDPG) and relabeled $B'$
		\EndFunction
		\end{algorithmic}
		\columnbreak
		\begin{algorithmic}
				\State \textbf{function} \textsc{Select }($*args$):\\
				\begin{adjustwidth}{\algorithmicindent}{0pt}
				Returns a behavioural goal for the agent.  Examples include the environment goal $g_\textrm{ext}$, a sample from the buffer of achieved goals $\mathcal{B}_{ag}$ \cite{warde-farley2018unsupervised}, or samples from a generative model such as a GAN \cite{florensa2018automatic} or VAE \cite{nair2018visual}. Our approach (MEGA) selects previously achieved goals in sparsely explored areas of the goal space according to a learned density model.
				\end{adjustwidth}
				\Statex \vspace{-2pt}
				\State \textbf{function} \textsc{Reward }($\Var{s}_t, \Var{a}_t, \Var{s}_{t+1}, \Var{g}$):\\
				\begin{adjustwidth}{\algorithmicindent}{0pt}
				Computes the environment reward or a learned reward function \cite{warde-farley2018unsupervised,nair2018visual}.
				\end{adjustwidth}
				\Statex \vspace{-2pt}
				\State \textbf{function} \textsc{Relabel }($B, *args$):\\
	        	\begin{adjustwidth}{\algorithmicindent}{0pt}
				Relabels goals and rewards in minibatch $B$ according to some strategy; e.g., don't relabel, \strat{future}, mix \strat{future} and generated goals \cite{nair2018visual}, or \strat{rfaab} (ours).
				\end{adjustwidth}
		\end{algorithmic}
		\end{multicols}\vspace{-0.05in}
	\end{small}
\end{algorithm*}

Despite the success of vanilla off-policy algorithms in dense-reward tasks, standard agents learn very slowly---or not at all---in sparse-reward, goal-conditioned tasks \cite{andrychowicz2017hindsight}. In order for a vanilla agent to obtain a positive reward signal and learn about goal $g$, the agent must stumble upon $g$ through random exploration \textit{while it is trying to achieve $g$}. Since the chance of this happening when exploring randomly decreases exponentially with the horizon (``the long horizon problem'') \cite{osband2014generalization}, successes are infrequent even for goals that are relatively close to the initial state, making learning difficult. 

One way to ameliorate the long horizon problem is based on the observation that,  regardless of the goal being pursued, $\langle$state, action, next state$\rangle$ transitions are unbiased samples from the environment dynamics. An agent is therefore free to pair transitions with any goal and corresponding reward, which allows it to use experiences gained in pursuit of one goal to learn about other goals (``goal relabeling'') \cite{kaelbling1993learning}. Hindsight Experience Replay (HER) \cite{andrychowicz2017hindsight} is a form of goal relabeling that relabels experiences with goals that are achieved later in the same trajectory. 
For every real experience, \citet{andrychowicz2017hindsight}'s \strat{future} strategy produces $k$ relabeled experiences, where the $k$ goals are sampled uniformly from goals achieved by future states in the same trajectory. This forms an implicit optimization curriculum, and allows an agent to learn about any goal $g$ it encounters during exploration. 

\begin{figure}[!b]
\centering
\vspace{-0.16in}
\includegraphics[width=2.2in]{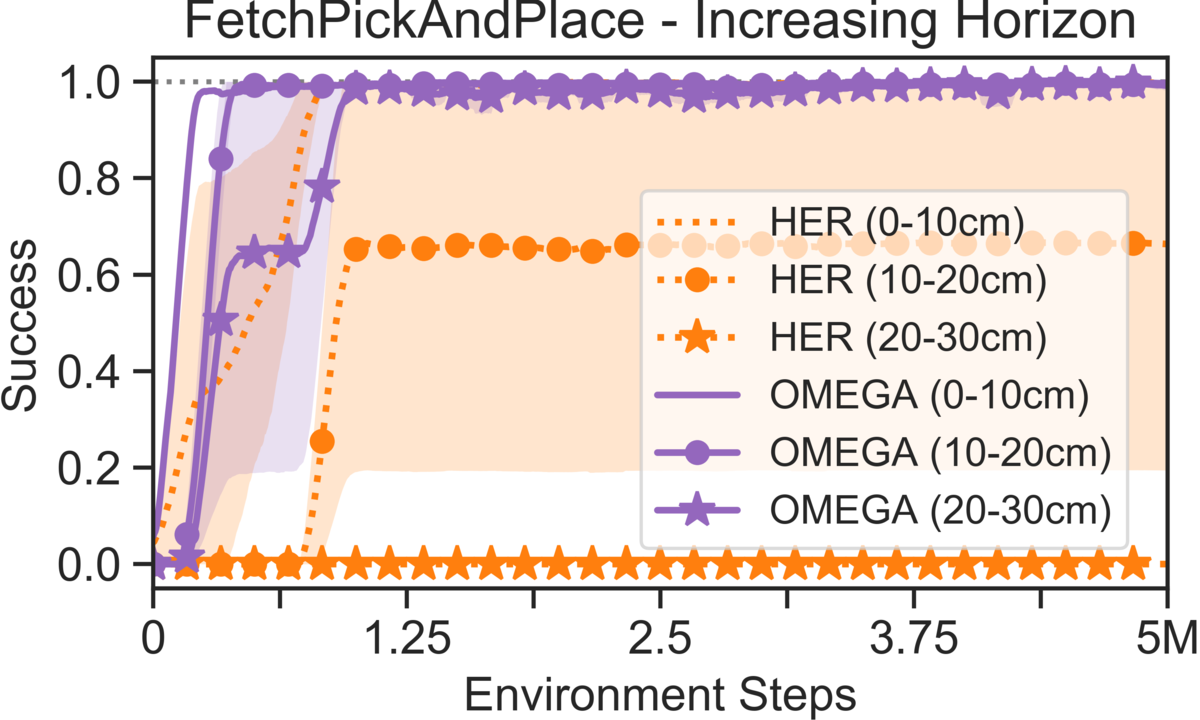}
\vspace{-0.1in}
\caption{Performance of a DDPG+HER agent that must lift a box to reach goals at increasing heights (3 seeds). As the horizon (desired height) increases, the agent loses the ability to solve the task in reasonable time. Our approach, OMEGA (Section \ref{section_method}), is robust to the horizon length. Specific details in Appendix.} \label{fig_her_horizon}
\end{figure}

Note, however, that a HER agent must still encounter $g$ (or goals sufficiently similar to $g$) in order to learn about $g$, and the long horizon problem persists for goals that are too far removed from the agent's initial state distribution. 
This is illustrated in Figure \ref{fig_her_horizon}, and is most easily understood by considering the tabular case, where no generalization occurs between a finite set of MDPs $\mathcal{M}_g$: since a learning signal is obtained only when transitioning into $s \in \success{g}$, the agent's achieved goal distribution must overlap with $\success{g}$ for learning to occur.
Empirically, this means that DDPG+HER agents that explore using only action noise or epsilon random actions fail to solve \textit{long-horizon tasks}, whose desired goal distribution does not overlap with the initial state distribution. This includes the original version of \env{FetchPickAndPlace} (with all goals in the air) \cite{andrychowicz2017hindsight}, block stacking \cite{nair2018overcoming}, and mazes \cite{trott2019keeping}.

\subsection{Setting intrinsic goals} \label{subsection_setting_intrinsic_goals}

We propose to approach the long-horizon problem by ignoring long-horizon goals: rather than try to achieve unobtainable goals, an agent can set its own intrinsic goals and slowly expand its domain of expertise in an unsupervised fashion. This is inspired by a number of recent papers on unsupervised multi-goal RL, to be described below. Our main contributions relative to past works are (1) a novel goal selection mechanism designed to address the long-horizon problem, and (2) a practical method to anneal initial unsupervised selection into training on the desired goals.

To capture the differences between various approaches, we present Algorithm \ref{alg_goal_seeking}, a unifying algorithmic framework for multi-goal agents. Variations occur in the subprocedures \textsc{Select}, \textsc{Reward}, and \textsc{Relabel}. The standard HER agent \citet{andrychowicz2017hindsight} \textsc{Selects} the environment goal $g_\textrm{ext}$, uses the environment \textsc{Reward} and uses the \strat{future} \textsc{Relabel} strategy. Functions used by other agents are detailed in Appendix \ref{appendix_agentcomparison}. We assume access to the environment \textsc{Reward} and propose a novel \textsc{Select} strategy---MaxEnt Goal Achievement (MEGA)---that initially samples goals from low-density regions of the achieved goal distribution. Our approach also leads to a novel  \textsc{Relabel} strategy, \strat{rfaab}, which samples from Real, Future, Actual, Achieved, and behavioural goals (detailed in Appendix \ref{appendix_implementation}).

Prior work also considers intrinsic \textsc{Select} functions. The approaches used by DISCERN \cite{warde-farley2018unsupervised}, RIG \cite{nair2018visual} and Skew-Fit \cite{pong2019skew} select goals using a model of the past achieved goal distribution. DISCERN samples from a replay buffer (a non-parametric model), whereas RIG and Skew-Fit learn and sample from a variational autoencoder (VAE) \cite{kingma2013auto}. These approaches are illustrated in Figure \ref{fig_density_based_select_comparison}, alongside HER and MEGA. Prior density-based approaches were not tailored to the long-horizon problem; e.g., DISCERN was primarily focused on learning an intrinsic \textsc{Reward} function, and left ``the incorporation of more explicitly instantiated [\textsc{Select}] curricula to
future work.'' By contrast, MEGA focuses on the low density, or sparsely explored, areas of the achieved goal distribution, forming a curriculum that crosses the gap between the initial state distribution and the desired goal distribution in record time.  Although Diverse sampling (e.g., Skew-Fit) is less biased towards already mastered areas of the goal space than Achieved sampling (e.g., RIG), we show in our experiments that it still under-explores relative to MEGA.

MEGA's focus on the frontier of the achieved goal set makes it similar to SAGG-RIAC \cite{baranes2013active}, which seeks goals that maximize learning progress, and Goal GAN \cite{florensa2018automatic}, which seeks goals of intermediate difficulty.

\section{Maximum Entropy Goal Achievement} \label{section_method}

\subsection{The MEGA and OMEGA objectives}

To motivate the MEGA objective, we frame exploration in episodic, multi-goal RL with goal relabeling as a distribution matching problem \cite{lee2019efficient}. We note that the original distribution matching objective is ill-conditioned in long-horizon problems, which suggests maximizing the entropy of the achieved goal distribution (the MEGA objective). We then show how this can be annealed into the original objective (the OMEGA objective). 

We start by noting that, for a truly off-policy agent, the actual goals used to produce the agent's experience do not matter, as the agent is free to relabel any experience with any goal. This implies that only the distribution of experience in the agent's replay buffer, along with the size of the buffer, matters for effective off-policy learning. How should an agent influence this distribution to accumulate useful data for achieving goals from the desired distribution $p_{dg}$? 

Though we lack a precise characterization of which data is useful, we know that all successful policies for goal $g$ pass through $g$, which suggests that useful data for achieving $g$ monotonically increases with the number of times $g$ is achieved during exploration.
Past empirical results, such as the success of \citet{andrychowicz2017hindsight}'s \strat{future} strategy and the effectiveness of adding expert demonstrations to the replay buffer \cite{nair2018overcoming}, support this intuition. Assuming a relatively fixed initial state distribution and uniformly distributed $p_{dg}$\footnote{For diverse initial state distributions, we would need to condition both $p_{dg}$ and $p_{ag}$ on the initial state. For non-uniform $p_{dg}$, we would likely want to soften the desired distribution as the marginal benefit of additional data is usually decreasing.}, 
it follows that the intrinsic goal $g^t$ at episode $t$ should be chosen to bring the agent's historical achieved goal distribution $p^{t}_{ag}$ closer to the desired distribution $p_{dg}$. We can formalize this as seeking $g^t$ to minimize the following distribution matching objective:
\begin{equation}\label{eq_og_distribution_matching}
 J_\textrm{original}(p^{t}_{ag}) = \DKL{p_{dg}}{p^{t}_{ag}},
\end{equation}
where $p^{t}_{ag}$ represents the \textit{historical} achieved goal distribution in the agent's replay buffer after executing its exploratory policy in pursuit of goal $g^t$. It is worth highlighting that objective (\ref{eq_og_distribution_matching}) is a forward KL: we seek $p_{ag}$ that ``covers'' $p_{dg}$ \cite{bishop2006pattern}. If reversed, it would always be infinite when $p_{dg}$ and the initial state distribution $s_0$ do not overlap, since $p_{dg}$ cannot cover $s_0$. 

So long as (\ref{eq_og_distribution_matching}) is finite and non-increasing over time, the support of $p_{ag}$ covers $p_{dg}$ and the agent is accumulating data that can be used to learn about all goals in the desired distribution. In those multi-goal environments where HER has been successful (e.g., \env{FetchPush}), this is easily achieved by setting the behavioural goal distribution $p_{bg}$ to equal $p_{dg}$ and using action space exploration \cite{plappert2018multi}. In long-horizon tasks, however, the objective (\ref{eq_og_distribution_matching}) is usually ill-conditioned (even undefined) at the beginning of training when the supports of $p_{dg}$ and $p_{ag}$ do not overlap. While this explains why HER with action space exploration fails in these tasks, it isn't very helpful, as the ill-conditioned objective is difficult to optimize.

When $p_{ag}$ does not cover $p_{dg}$, a natural objective is to expand the support of $p_{ag}$, in order to make the objective (\ref{eq_og_distribution_matching}) finite as fast as possible. 
We often lack a useful inductive bias about which direction to expand the support in; e.g., a naive heuristic like Euclidean distance in feature space can be misleading due to dead-ends or teleportation \cite{trott2019keeping}, and should not be relied on for exploration. In absence of a useful inductive bias, it is sensible to expand the support as fast as possible, in any and all directions as in breadth-first search, which can be done by maximizing the entropy of the achieved goal distribution $H[p_{ag}]$. We call this the Maximum Entropy Goal Achievement (MEGA) objective:
\begin{equation}\label{eq_obj_entropy_max}
J_\textrm{MEGA}(p^{t}_{ag}) = -H[p^{t}_{ag}],
\end{equation}
The hope is that by maximizing $H[p_{ag}]$ (minimizing $J_\textrm{MEGA}$), the agent will follow a natural curriculum, expanding the size of its achievable goal set until it covers the support of the desired distribution $p_{dg}$ and objective (\ref{eq_og_distribution_matching}) becomes tractable. 

In the unsupervised case, where $p_{dg}$ is not specified, the agent can stop at the MEGA objective. 
In the supervised case we would like the agent to somehow anneal objective (\ref{eq_obj_entropy_max}) into objective
(\ref{eq_og_distribution_matching}). We can do this by approximating (\ref{eq_obj_entropy_max}) using a distribution matching objective, where the desired distribution is uniform over the current support:
\begin{equation}\label{eq_ag_distribution_matching}\small
	\tilde J_\textrm{MEGA}(p^{t}_{ag}) = \DKL{\mathcal{U}(\supp(p^{t}_{ag}))}{p^{t}_{ag}}.
\end{equation}
This is a sensible approximation, as it shares a maximum with (\ref{eq_obj_entropy_max}) when the uniform distribution over $G$ is obtainable, and encourages the agent to ``cover'' the current support of the achieved goal distribution as broadly as possible, so that the diffusion caused by action space exploration will increase entropy. We may now form the mixture distribution $p^t_\alpha = \alpha p_{dg} + (1-\alpha) \mathcal{U}(\supp(p^t_{ag}))$ and state our final ``OMEGA'' objective, which anneals the approximated MEGA into the original objective:
\begin{equation}\label{eq_final_distribution_matching}\small
	J_\textrm{OMEGA}(p^{t}_{ag}) = \DKL{p_\alpha}{p^{t}_{ag}}.
\end{equation}
The last remaining question is, how do we choose $\alpha$? We would like $\alpha = 0$ when $p_{ag}$ and $p_{dg}$ are disconnected, and $\alpha$ close to $1$ when $p_{ag}$ well approximates $p_{dg}$. One way to achieve this, which we adopt in our experiments, is to set $$\alpha = 1/\max(b + \DKL{p_{dg}}{p_{ag}}, 1),$$ where $b \leq 1$. The divergence is infinite ($\alpha = 0$) when $p_{ag}$ does not cover $p_{dg}$ and approaches $0$ ($\alpha = 1$) as $p_{ag}$ approaches $p_{dg}$. Our experiments use $b = -3$, which we found sufficient to ensure $\alpha=1$ at convergence (with $b = 1$, we may never have $\alpha = 1$, since $p_{ag}$ is historical and biased towards the initial state distribution $s_0$). 

\subsection{Optimizing the MEGA objective} \label{subsection_optimizing_mega}

We now consider choosing behavioural goal $\hat g \sim p_{bg}$ in order to optimize the MEGA objective (\ref{eq_obj_entropy_max}), as it is the critical component of (\ref{eq_final_distribution_matching}) for early exploration in long-horizon tasks and general unsupervised goal-seeking. In supervised tasks, the OMEGA objective (\ref{eq_final_distribution_matching}) can be approximately optimized by instead using the environment goal with probability $\alpha$.

We first consider what behavioural goals we would pick if we had an oracle that could predict the conditional distribution $q(g'\given\hat g)$ of goals $g'$ that would be achieved by conditioning the policy on $\hat g$. Then, noting that this may be too difficult to approximate in practice, we propose a minimum density heuristic that performs well empirically. 
The resulting \textsc{Select} functions are shown in Algorithm \ref{alg_mega_select}.

\paragraph{Oracle strategy}  If we knew the conditional distribution $q(g'\given\hat g)$ of goals $g'$ that would be achieved by conditioning behaviour on $\hat g$, we could compute the expected next step MEGA objective as the expected entropy of the new empirical 
$p_{ag \given g'}$
after sampling $g'$ and adding it to our buffer:
\begin{align*}\small
J_\textrm{MEGA}(p_{ag \given g'}) &= -\mathbb{E}_{g' \sim q(g' \given \hat g)} H[p_{ag \given g'}] \\
&\hspace{-0.4in}= \sum_{g'} q(g'|\hat{g}) \sum_{g} p_{ag \given g'}(g) \log p_{ag \given g'}(g),
\end{align*}
To explicitly compute this objective one must compute both the new distribution and its entropy for each possible new achieved goal $g'$. 
The following result simplifies matters in the tabular case. Proofs may be found in Appendix \ref{appendix_props}. 

\begin{proposition}[Discrete Entropy Gain]
Given buffer $\mathcal{B}$ with $\eta\!=\!\frac{1}{|\mathcal{B}|}$, maximizing expected next step entropy is equivalent to maximizing expected point-wise entropy gain $\Delta H (g')$:
\begin{equation}\label{eq_prop_entropy_gain}
\begin{split}\small
 \hat g^* &=\arg\max_{\hat{g} \in \mathcal{B}} \mathbb{E}_{g' \sim q(g' \given \hat g)} H[p_{ag \given g'}] \\
 &= \arg\max_{\hat{g}\in \mathcal{B}} \mathbb{E}_{g' \sim q(g' \given \hat g)} \Delta H (g'),
\end{split}
\end{equation}
where $\Delta H (g') = p_{ag}(g') \log p_{ag}(g')\ - $\\[2pt]
\hspace*{1mm}\hfill$(p_{ag}(g') + \eta) \log (p_{ag}(g') + \eta)$.
\end{proposition}

For most agents $\eta$ will quickly approach $0$ as they accumulate experience, so that choosing $\hat g$ according to (\ref{eq_prop_entropy_gain}) becomes equal (in the limit) to choosing $\hat g$ to maximize the directional derivative $\langle  \nabla_{p_{ag}} H[p_{ag}], q(g' \given \hat g) - p_{ag}\rangle$.

\begin{proposition}[Discrete Entropy Gradient]\label{prop_discrete_entropy_gradient}
\begin{equation}\small
\begin{split}
    \lim_{\eta \to 0} \hat g^* &= \arg\max_{\hat{g}\in \mathcal{B}}\langle \nabla_{p_{ag}} H[p_{ag}], q(g' \given \hat g) - p_{ag} \rangle\\
    &= \arg\max_{\hat{g}\in \mathcal{B}} \DKL{q(g' \given \hat g)}{p_{ag}} + H[q(g' \given \hat g)]
\end{split}%
\end{equation}%
\end{proposition}%
This provides a nice intuition behind entropy gain exploration: we seek maximally diverse outcomes ($H[q(g' \given \hat g)]$) that are maximally different from historical experiences ($\DKL{q(g' \given \hat g)}{p_{ag}}$)---i.e., exploratory behavior should evenly explore under-explored regions of the state space. 
By choosing goals to maximize the entropy gain, an agent effectively performs constrained gradient ascent \cite{frank1956algorithm,hazan2018provably} on the entropy objective.

Assuming the empirical $p_{ag}$ is used to induce (abusing notation) a density $p_{ag}$ with full support, Proposition \ref{prop_discrete_entropy_gradient} extends to the continuous case by taking the functional derivative of the differential entropy with respect to the variation $\eta(g) = q(g' \given \hat g)(g) - p_{ag}(g)$ (Appendix \ref{appendix_props}).

\algrenewcommand\algorithmiccomment[1]{\hfill #1}

\begin{algorithm}[t]
	\caption{O/MEGA \textsc{Select} functions} \label{alg_mega_select}
	\begin{small}
		\begin{algorithmic}
		\Function{OMEGA\_Select }{env goal $g_\textrm{ext}$, bias $b$, $*args$}:
		\State $\alpha \gets 1/\max(b + \DKL{p_{dg}}{p_{ag}}, 1)$
		\If{$x \sim \mathcal{U}(0, 1) < \alpha$} \Return $g_\textrm{ext}$
		\Else \ \Return \textsc{MEGA\_Select}($*args$)
		\EndIf
		\EndFunction
		\Statex
		\Function{MEGA\_Select }{buffer $\mathcal{B}$, num\_candidates $N$}:
		\State Sample $N$ candidates $\{g_i\}_{i=1}^N \sim \mathcal{B}$
	    \State Eliminate unachievable candidates (see text)
		\State \Return $\hat g = \argmin_{g_i} \hat p_{ag}(g_i)$ \Comment{$(*)$}
		\EndFunction
		\end{algorithmic}
	\end{small}
\end{algorithm}

\paragraph{Minimum density approximation} Because we do not know $q(g'\given\hat g)$, we must approximate it with either a learned model or an effective heuristic. The former solution is difficult, because by the time there is enough data to make an accurate prediction conditional on $\hat g$, $q(g'\given\hat g)$ will no longer represent a sparsely explored area of the goal space. 
While it might be possible to make accurate few- or zero-shot predictions if an agent accumulates enough data in a long-lived, continual learning setting with sufficient diversity for meta-learning \cite{ren2018meta}, in our experiments we find that a simple, minimum-density approximation, which selects goals that have minimum density according to a learned density model, is at least as effective (Appendix \ref{appendix_additional_experiments}). 
We can view this approximation as a special case where the conditional $q(g'\given\hat g) = \mathds{1}[g' = \hat g]$, i.e. that the agent achieves the behaviour goal.

\begin{proposition}\label{prop_det_cond_approx}
If $q(g'| \hat g) = \mathds{1}[g' = \hat g]$, the discrete entropy gradient objective simplifies to a minimum density objective:
\begin{equation}\small
\begin{split}
    \hat g^* 
    &= \arg\max_{\hat{g}\in \mathcal{B}} -\log [p_{ag}(\hat g)] \\
    &= \arg\min_{\hat{g}\in \mathcal{B}} p_{ag}(\hat g).
\end{split}%
\end{equation}%
\end{proposition}%
Our minimum density heuristic (Algorithm \ref{alg_mega_select}) fits a density model to the achieved goals in the buffer to form estimate $\hat p_{ag}$ of the historical achieved goal distribution $p_{ag}$ and uses a generate and test strategy \cite{newell1969artificial} that samples $N$ candidate goals $\{g_i\}_{i=1}^N \sim \mathcal{B}$ from the achieved goal buffer (we use $N=100$ in our experiments) and selects the minimum density candidate $\hat g = \argmin_{g_i} \hat p_{ag}(g_i)$. We then adopt a Go Explore \cite{ecoffet2019go} style strategy, where the agent increases its action space exploration once a goal is achieved. Intuitively, this heuristic seeks out past achieved goals in sparsely explored areas, and explores around them, pushing the frontier of achieved goals forward. 

It is important for $\hat{g}$ to be achievable. If it is not, then $q(g'\given\hat g)$ may be disconnected from $\hat{g}$, as is the case when the agent pursues unobtainable $g_\textrm{ext}$ (Figure \ref{fig_her_horizon}), which undermines the purpose of the minimum density heuristic. To promote achievability, our experiments make use of two different mechanisms. First, we only sample candidate goals from the past achieved goal buffer $\mathcal{B}$. Second, we eliminate candidates whose estimated value (according to the agent's goal-conditioned Q-function) falls below a dynamic cutoff, which is set according to agent's goal achievement percentage during exploration. The specifics of this cutoff mechanism may be found in Appendix \ref{appendix_implementation}. Neither of these heuristics are core to our algorithm, and they might be be replaced with, e.g., a generative model designed to generate achievable goals \cite{florensa2018automatic}, or a success predictor that eliminates unachievable candidates.

\section{Experiments} \label{section_empirical}

Having described our objectives and proposed approaches for optimizing them, we turn to evaluating our O/MEGA agents on four challenging, long-horizon environments 
that standard DDPG+HER agents fail to solve. We compare the performance of our algorithms with several goal selection baselines. To gain intuition on our method, we visualize qualitatively the behaviour goals selected and quantitatively the estimated entropy of the achieved goal distribution. 

\begin{figure*}[!t]
    \begin{subfigure}[b]{0.24\textwidth}
         \centering
         \includegraphics[width=\textwidth]{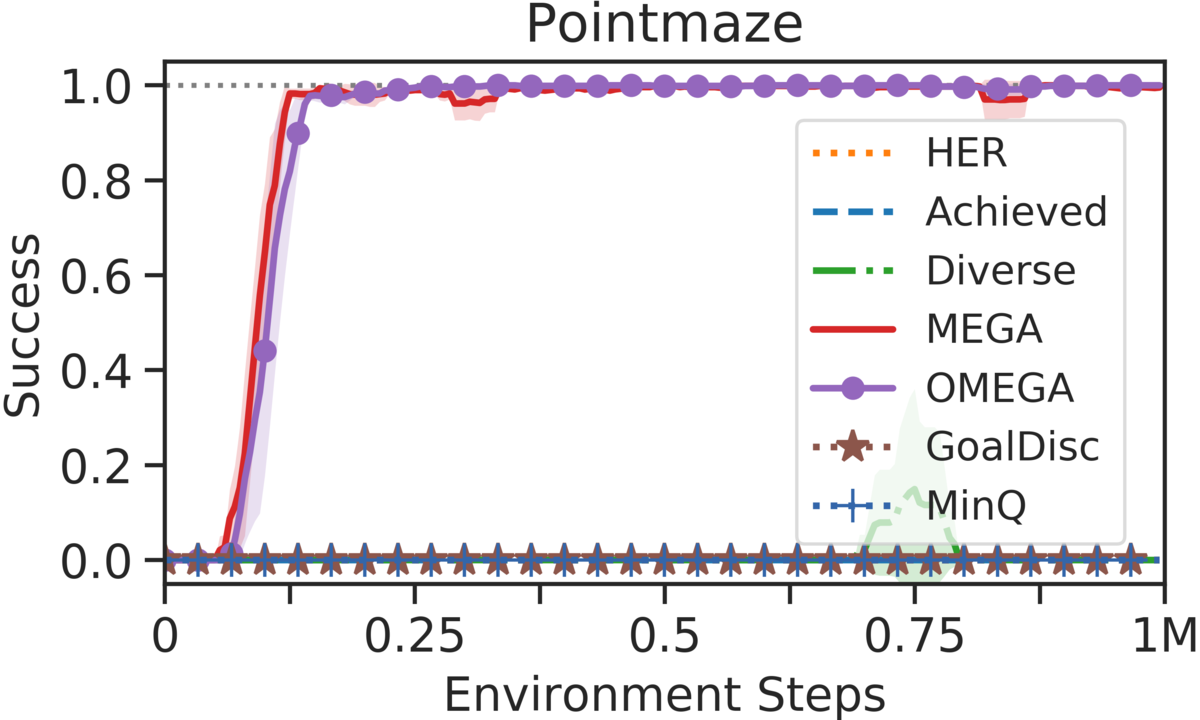}
         \label{fig:}
     \end{subfigure}
     \hfill
     \begin{subfigure}[b]{0.24\textwidth}
         \centering
         \includegraphics[width=\textwidth]{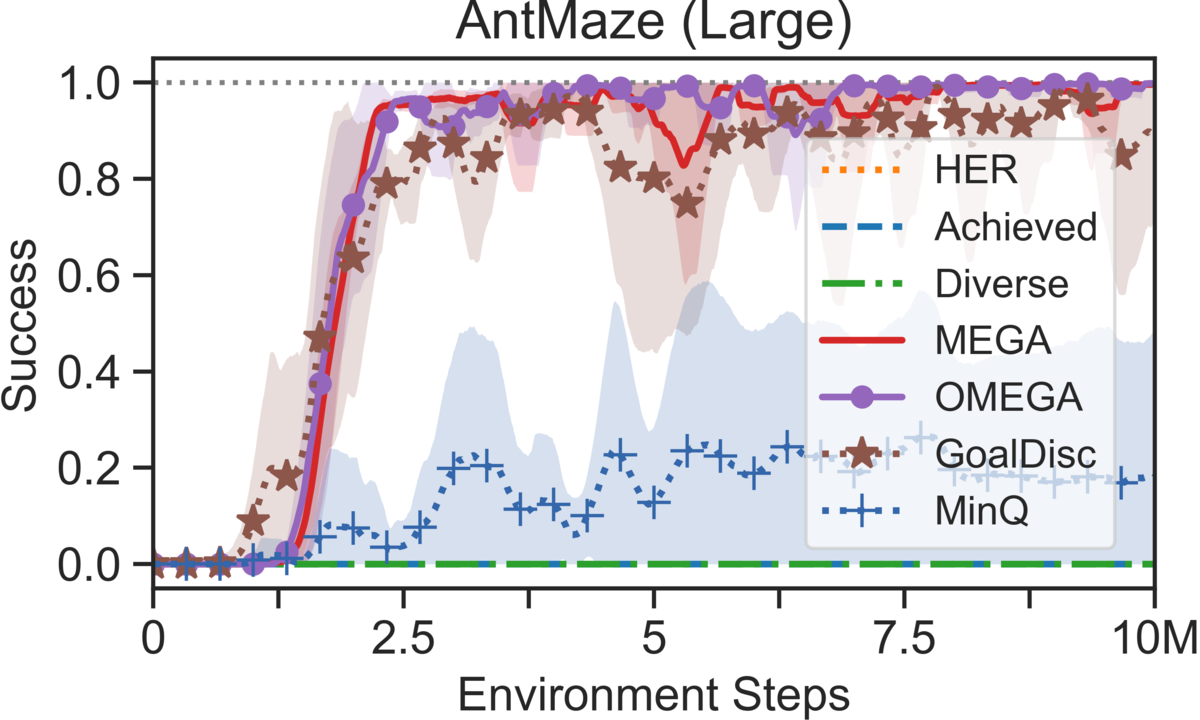}
         \label{fig:}
     \end{subfigure}
     \hfill
     \begin{subfigure}[b]{0.24\textwidth}
         \centering
         \includegraphics[width=\textwidth]{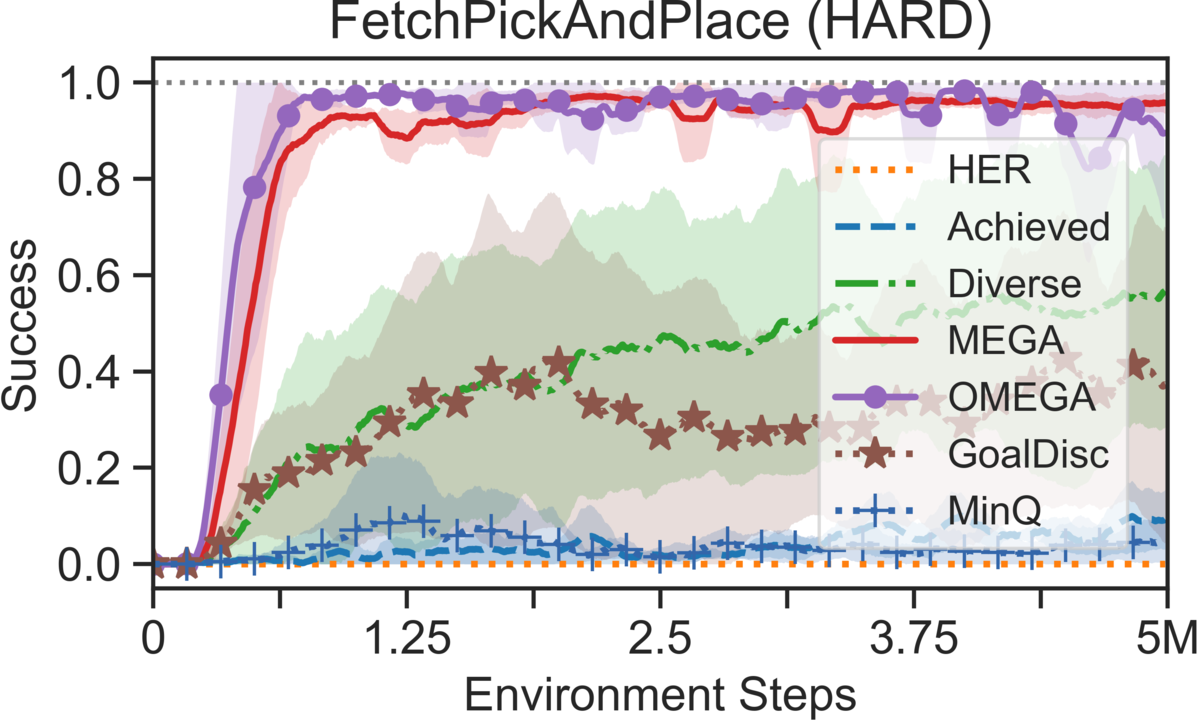}
         \label{fig:}
     \end{subfigure}
     \hfill
     \begin{subfigure}[b]{0.24\textwidth}
         \centering
         \includegraphics[width=\textwidth]{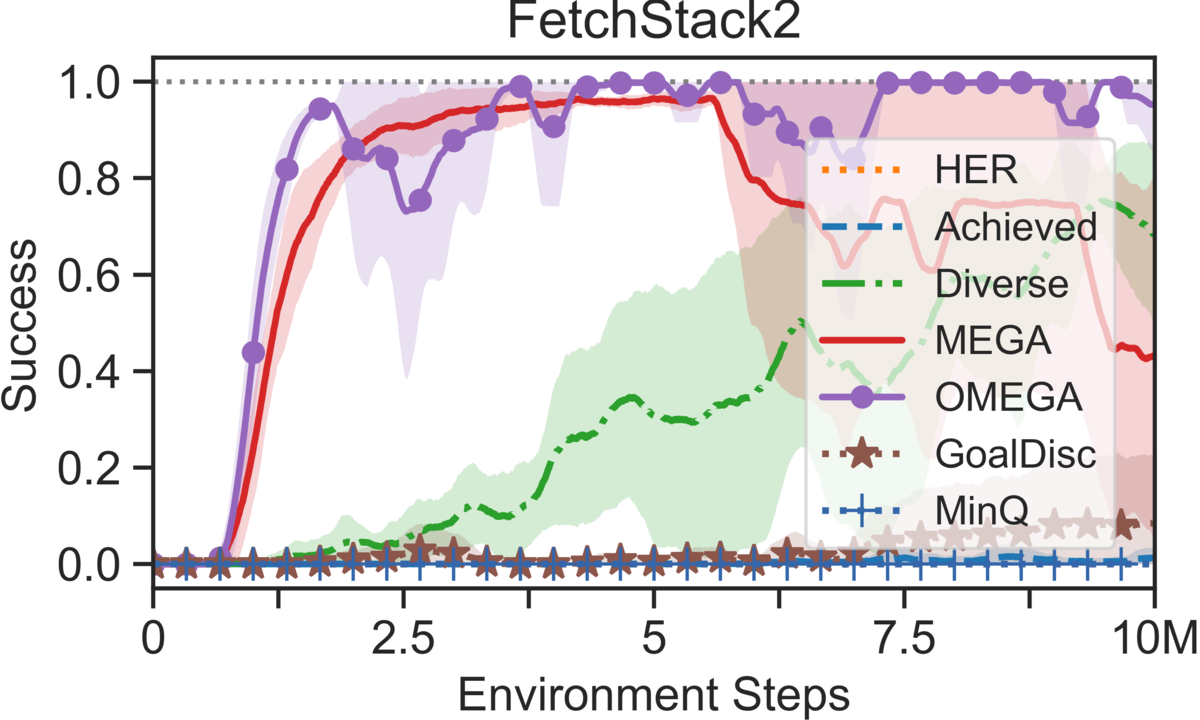}
         \label{fig:}
     \end{subfigure}
     \vspace{-0.3in}
     \caption{Test success on the desired goal distribution, evaluated throughout training, for several behaviour goal selection methods (3 seeds each). Our methods (MEGA and OMEGA) are the only the methods which are able to solve the tasks with highest sample efficiency. In \env{FetchStack2} we see that OMEGA's eventual focus on the desired goal distribution is necessary for long run stability.}
    \label{fig:main_results}
\end{figure*}

\begin{figure*}[t]
     \centering
    \includegraphics[width=\textwidth]{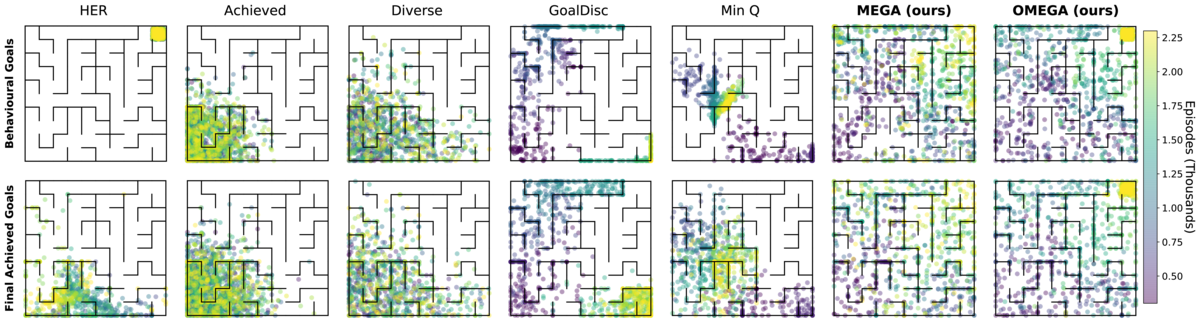}
    \vspace{-0.3in} 
     \caption{Visualization of behavioural (\textbf{top}) and terminal achieved (\textbf{bottom}) goals in \env{PointMaze}, colour-coded for over the course of training for several behavioural goal sampling methods. Only our methods reach the desired goal area in top right hand corner in approximately 2000 episodes, beating the previous state of the art \cite{trott2019keeping} by almost 2 orders of magnitude (100 times). }
    \vspace{-0.1in} 
    \label{fig:goal_vis}
\end{figure*}

\paragraph{Environments}

We consider four environments. In \env{PointMaze} \cite{trott2019keeping}, a point must navigate a 2d maze, from the bottom left corner to the top right corner. In \env{AntMaze} \cite{nachum2018data,trott2019keeping}, an ant must navigate a U-shaped hallway to reach the target. In \env{FetchPickAndPlace} (hard version) \cite{plappert2018multi}, a robotic arm must grasp a box and move it to the desired location that is at least 20cm in the air. In \env{FetchStack2} \cite{nair2018overcoming}, a robotic arm must move each of the two blocks into the desired position, where one of the block rests on top of the other. 
In \env{PointMaze} and \env{AntMaze} goals are 2-dimensional and the agent is successful if it reaches the goal once. In \env{FetchPickAndPlace} and \env{FetchStack2}, goals are 3- and 6-dimensional, respectively, and the agent must maintain success until the end of the episode for it to count. 

\paragraph{Baselines} We compare MEGA and OMEGA to the three density-based \textsc{Select} mechanisms shown in Figure \ref{fig_density_based_select_comparison} above: sampling from $p_{dg}$ (``HER''), sampling from the historical $p_{ag}$ as done approximately by RIG (``Achieved''), and sampling from a skewed historical $p_{ag}$ that is approximately uniform on its support, as done by DISCERN and Skew-Fit (``Diverse''). We also compare against non density-based baselines as follows. \env{PointMaze} and \env{AntMaze} are the same environments used by the recent Sibling Rivalry paper \cite{trott2019keeping}. Thus, our results are directly comparable to Figure 3 of their paper, which tested four algorithms: HER, PPO \cite{schulman2017proximal}, PPO with intrinsic curiosity \cite{pathak2017curiosity}, and PPO with Sibling Rivalry (PPO+SR). The \env{AntMaze} environment uses the same simulation as the \env{MazeAnt} environment tested in the Goal GAN paper \cite{florensa2018automatic}, but is four times larger. In Appendix \ref{appendix_additional_experiments}, we test MEGA on the smaller maze and obtain an almost 1000x increase in sample efficiency as compared to Goal GAN and the Goal GAN implementation of SAGG-RIAC \cite{baranes2013active}. Results are not directly comparable as Goal GAN uses an on-policy TRPO base \cite{schulman2015trust}, which is very sample inefficient relative to our off-policy DDPG base. Thus, we adapt the Goal GAN discriminator to our setting by training a success predictor to identify goals of intermediate difficulty (Appendix \ref{appendix_implementation}) (``GoalDisc''). Finally, we compare against a minimum Q heuristic, which selects distant goals \cite{hartikainen2020dynamical} (``MinQ'').

We note a few things before moving on. First, Sibling Rivalry \cite{trott2019keeping} is the only prior work that directly addresses the long-horizon, sparse reward problem (without imitation learning). Other baselines were motivated by and tested on other problems. Second, the generative parts of Goal GAN and RIG are orthogonal to our work, and could be combined with MEGA-style generate-and-test selection, as we noted above in Section \ref{subsection_optimizing_mega}. We adopted the generative mechanism of DISCERN (sampling from a buffer) as it is simple and has a built-in bias toward achievable samples. For a fair comparison, all of our implemented baselines use the same buffer-based generative model and benefit from our base DDPG+HER implementation (Appendix \ref{appendix_implementation}). The key difference between MEGA and our implemented baselines is the \textsc{Select} mechanism (line $(*)$ of Algorithm \ref{alg_mega_select}). 

\paragraph{Main results}

Our main results, shown in Figure \ref{fig:main_results} clearly demonstrate the advantage of minimum density sampling. We confirm that desired goal sampling (HER) is unable to solve the tasks, and observe that Achieved and Diverse goal sampling fail to place enough focus on the frontier of the achieved goal distribution to bridge the gap between the initial state and desired goal distributions. On \env{PointMaze}, none of the baselines were able to solve the environment within 1 million steps. The best performing algorithm from \citet{trott2019keeping} is PPO+SR, which solves \env{PointMaze} to 90\% success in approximately 7.5 million time steps (O/MEGA is almost 100 times faster). On \env{AntMaze}, only MEGA, OMEGA and the GoalDisc are able to solve the environment. The best performing algorithm from \citet{trott2019keeping} is hierarchical PPO+SR, which solves \env{AntMaze} to 90\% success in approximately 25 million time steps (O/MEGA is roughly 10 times faster). On a maze that is four times smaller, \citet{florensa2018automatic} tested four algorithms, including SAGG-RIAC \cite{baranes2013active}, which was implemented, along with Goal GAN, using a TRPO base. Their best performing result achieves 71\% coverage of the maze in about 175 million time steps (O/MEGA is roughly 100 times faster on a larger maze). O/MEGA also demonstrates that \env{FetchStack2} can be solved from scratch, without expert demonstrations \cite{duan2017one,nair2018overcoming} or a task curriculum \cite{colas2018curious}.

\begin{figure}[t]
    \begin{subfigure}[b]{0.23\textwidth}
         \centering
         \includegraphics[width=\textwidth]{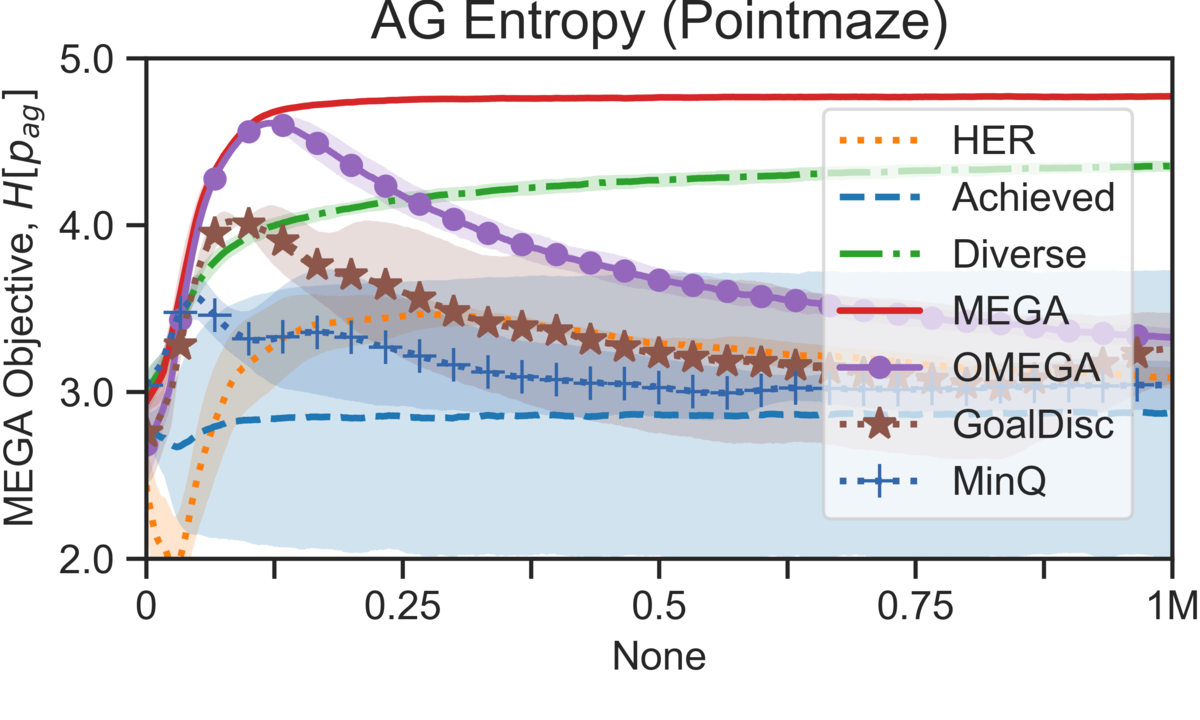}
     \end{subfigure}
     \hfill
     \begin{subfigure}[b]{0.23\textwidth}
         \centering
         \includegraphics[width=\textwidth]{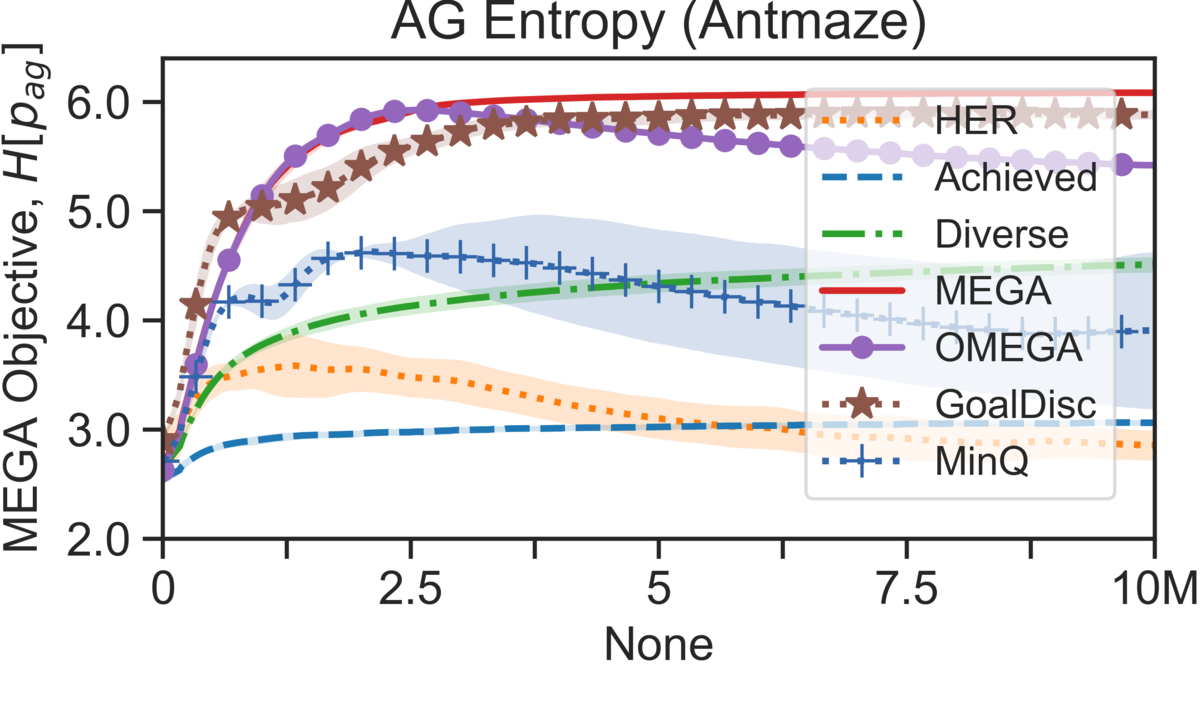}
     \end{subfigure}
    \begin{subfigure}[b]{0.23\textwidth}
         \centering
         \includegraphics[width=\textwidth]{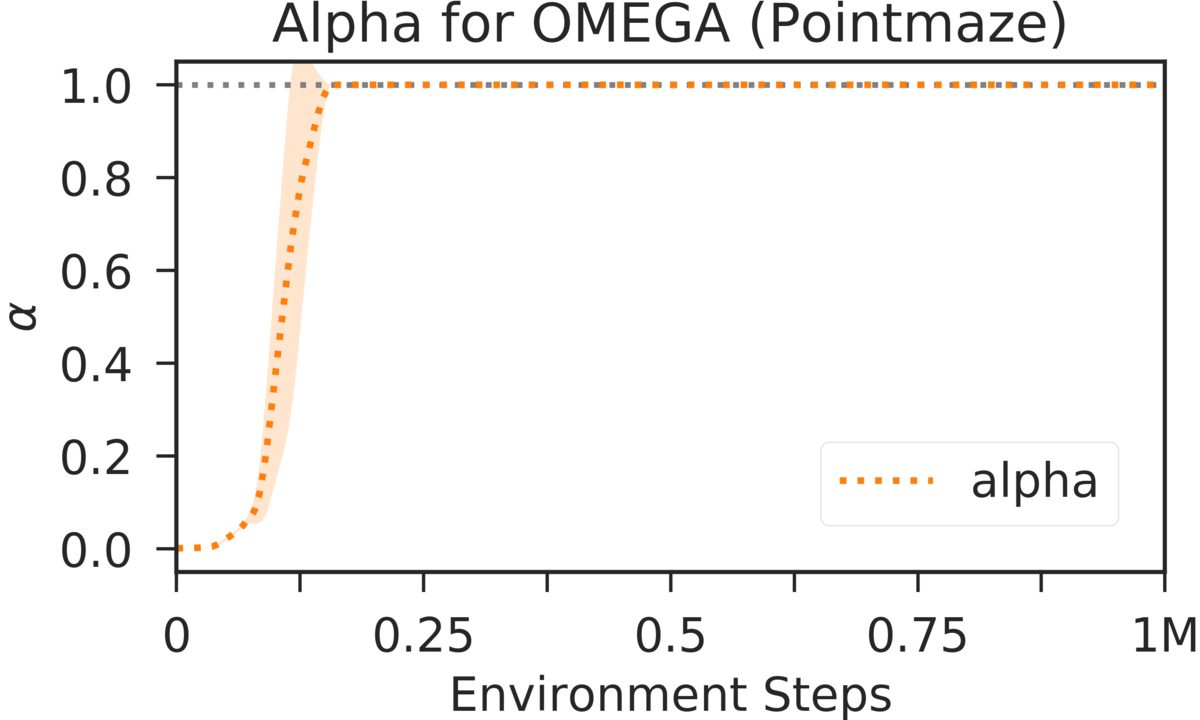}
     \end{subfigure}
     \hfill
     \begin{subfigure}[b]{0.23\textwidth}
         \centering
         \includegraphics[width=\textwidth]{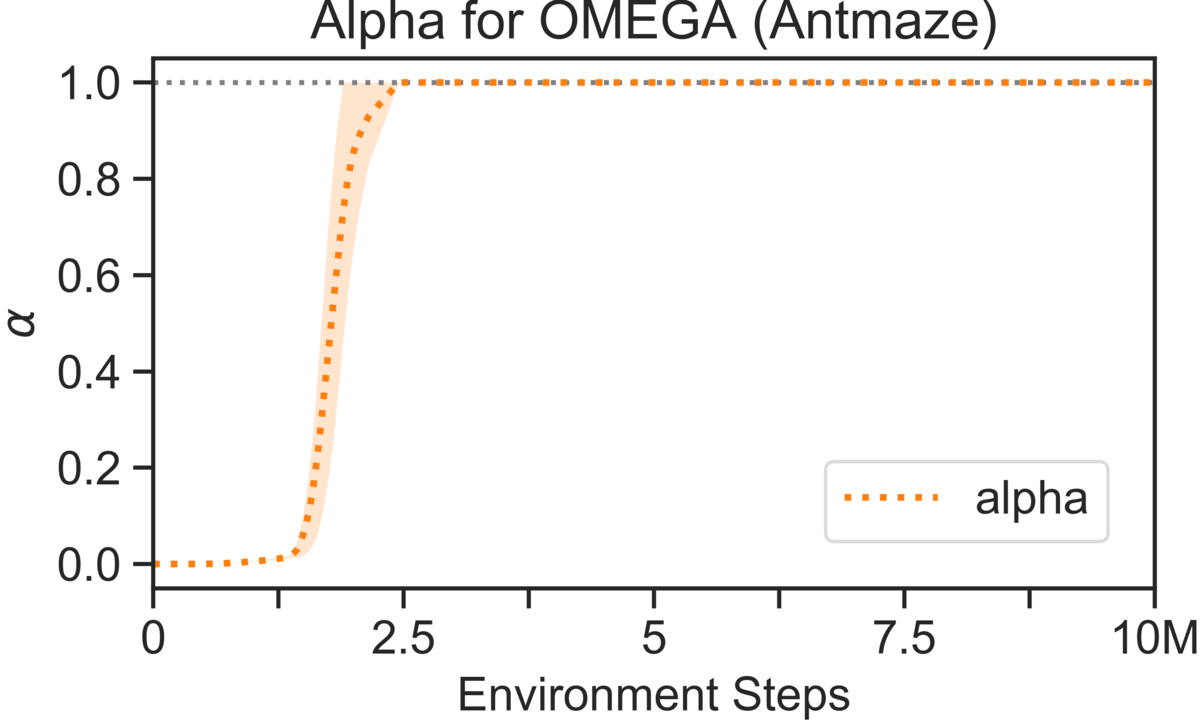}
     \end{subfigure}
    \vspace{-0.1in}
    \caption{\textbf{Top}: Entropy of the achieved goal buffer for \texttt{Pointmaze} (left) and \texttt{Antmaze} (right) over course of training, estimated using a Kernel Density Estimator. O/MEGA expand the entropy much faster than the baselines. \textbf{Bottom}: $\alpha$ computed by OMEGA, which transitions from intrinsic to extrinsic goals.
    }
    \label{fig:main_kde_entropy}
\end{figure}

\paragraph{Maximizing entropy} In Figure \ref{fig:main_kde_entropy} (top), we observe that our approach increases the empirical entropy of the achieved goal buffer (the MEGA objective) much faster than other goal sampling methods. MEGA and OMEGA rapidly increase the entropy and begin to succeed with respect to the desired goals as the maximum entropy is reached. As OMEGA begins to shift towards sampling mainly from the desired goal distribution (Figure \ref{fig:main_kde_entropy} (bottom)), the entropy declines as desired goal trajectories become over represented. We observe that the intermediate difficulty heuristic (GoalDisc) is a good optimizer of the MEGA objective on \env{AntMaze}, likely due to the environment's linear structure. This explains its comparable performance to MEGA.

\begin{figure*}[t]
     \centering
    \includegraphics[width=\textwidth]{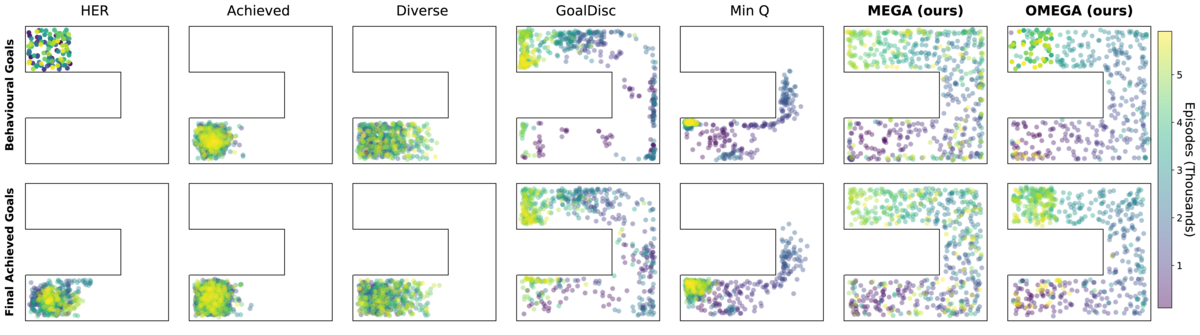}
    \vspace{-0.3in} 
     \caption{Visualization of behavioural (\textbf{top}) and terminal achieved (\textbf{bottom}) goals in \env{AntMaze}, colour-coded for over the course of training for several behavioural goal sampling methods. The only baseline that reached the desired goal in the top left was GoalDisc. }
    \vspace{-0.1in} 
    \label{fig:goal_vis_antmaze_main}
\end{figure*}

\paragraph{Visualization of achieved goals}

To gain intuition for how our method compares to the baselines, we visualize the terminal achieved goal at the end of the episodes throughout the training for \env{PointMaze} in Figure \ref{fig:goal_vis}. Both MEGA and OMEGA set goals that spread outward from the starting location as training progresses, akin to a breadth-first search, with OMEGA eventually transitioning to goals from the desired goal distribution in the top right corner. Diverse sampling maintains a fairly uniform distribution at each iteration, but explores slowly as most goals are sampled from the interior of the support instead of the frontier. Achieved sampling oversamples goals near the starting location and suffers from a ``rich get richer'' problem. Difficulty-based GoalDisc and distance-based MinQ sampling explore deeply in certain directions, akin to a depth-first search, but ignore easier/closer goals that can uncover new paths. 

A similar visualization for \env{AntMaze} is shown in Figure \ref{fig:goal_vis_antmaze_main}. Aside from our methods, the only baseline able to reach the desired goal area is GoalDisc. MEGA and OMEGA observe a higher diversity in achieved goals, which suggests the learned policy from our methods will be more robust than the GoalDisc policy if the desired goal distribution changes, but we did not directly test this hypothesis. 

\section{Other Related Work} \label{section_related}

\begin{figure}[t]
    \begin{subfigure}[b]{0.23\textwidth}
         \centering
         \includegraphics[width=\textwidth]{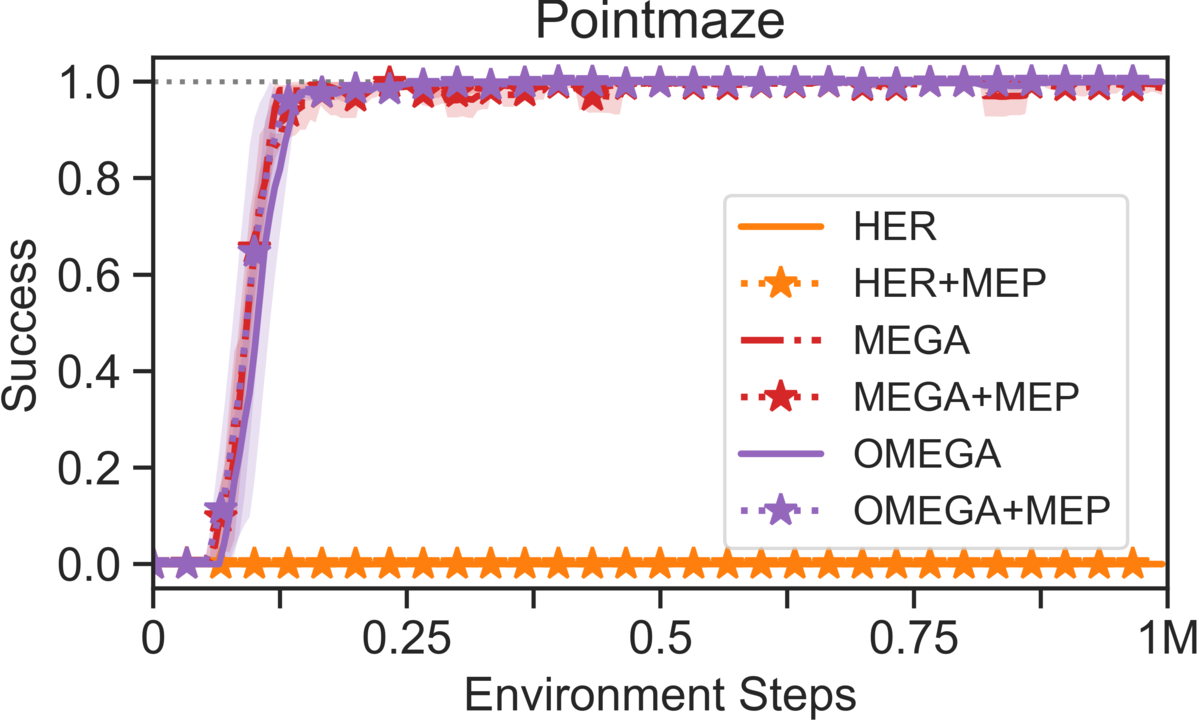}
     \end{subfigure}
     \begin{subfigure}[b]{0.23\textwidth}
         \centering
         \includegraphics[width=\textwidth]{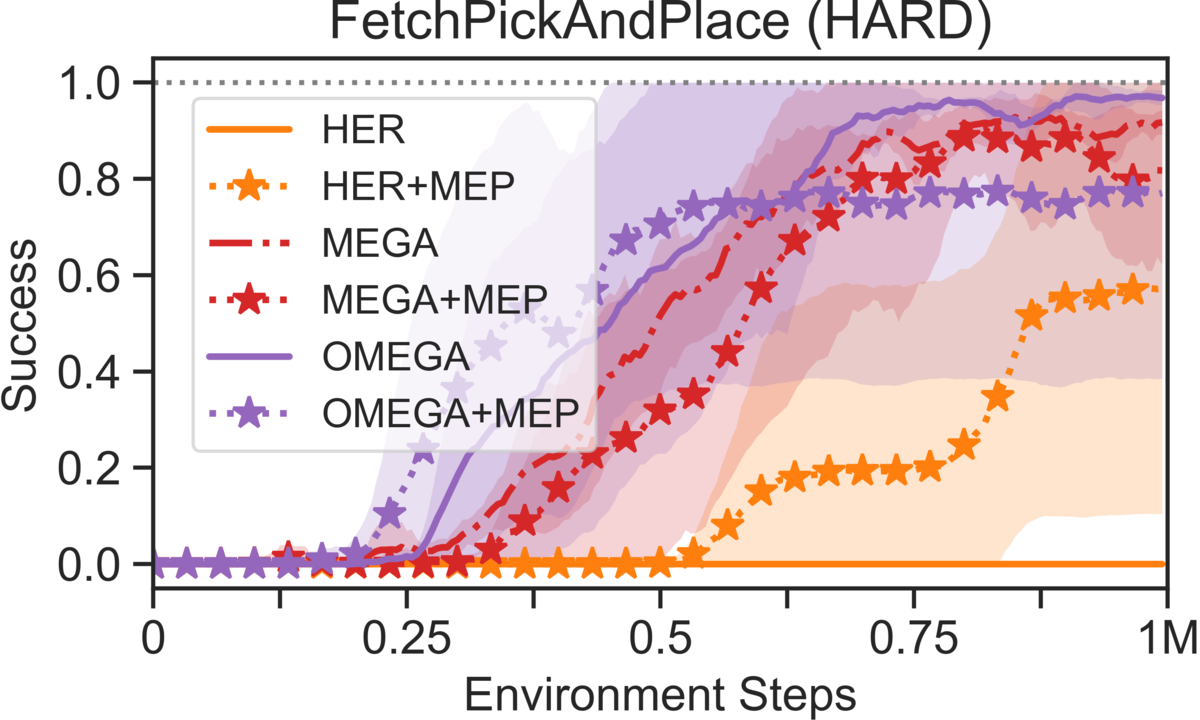}
     \end{subfigure}
    \vspace{-0.1in}
    \caption{MEP \cite{zhao2019maximum} maximizes the entropy of training goals in the \textsc{Optimize} method.
    While MEP can help the function approximator generalize, and allows HER to achieve some success in \env{FetchPickAndPlace} (hard), it does not directly help the agent explore and cross long horizon gaps. 
    }
    \label{fig:mep}
\end{figure}

\paragraph{Maximum entropy-based prioritization (MEP)}  \label{related_mep} While MEGA influences the entropy of the historical achieved goal distribution during \textsc{Select}, MEP \cite{zhao2019maximum} reweighs experiences during \textsc{Optimize} to increase the entropy of the goals in an agent's training distribution. Unlike MEGA, MEP does not set intrinsic goals and does not directly influence an agent's exploratory behavior. As a result, MEP is limited to the support set of the observed achieved goals and must rely on the generalization of the neural network model to cross long-horizon gaps. 
As MEGA and MEP can be applied simultaneously, we compared using HER and O/MEGA, with and without MEP in \env{PointMaze} and \env{FetchPickAndPlace}. As shown in Figure \ref{fig:mep}, applying MEP to HER helps the agent achieved some success in the \env{FetchPickAndPlace}, but is unable to help in the \env{PointMaze} where the long horizon gap is more severe. Combining MEGA and MEP has limited effect, possibly because a MEGA agent's achieved goal distribution is already close to uniform. 
See Appendix \ref{appendix_implementation} (``MEP'') for details.

\paragraph{Curiosity} Maximizing entropy in the goal space is closely related to general RL (not multi-goal) algorithms that seek to maximize entropy in the state space \cite{hazan2018provably, lee2019efficient} or grant the agent additional reward based on some measure of novelty, surprise or learning progress \cite{kolter2009near,schmidhuber2010formal,lopes2012exploration,bellemare2016unifying,ostrovski2017count,tang2017exploration,pathak2017curiosity,burda2018exploration}. Two key differences should be noted. First, MEGA uses a low-dimensional, abstract goal space to drive exploration in meaningful directions \cite{baranes2010intrinsically}. This focuses the agent on what matters, and avoids the ``noisy-TV'' problem \cite{burda2018exploration}. As this requires a known, semantically meaningful goal space, future work might explore how one can automatically choose a good goal space \cite{lee2020weakly}. Second, MEGA agents learn and use a goal-conditioned policy, which makes MEGA exploration more ``active'' than exploration based on intrinsic reward  \cite{shyam2018model}. It is reasonable to interpret the low density region of an agent's achievable goal space as its ``frontier'', so that MEGA exploration is a form of \textit{frontier exploration} \cite{yamauchi1997frontier,holz2010evaluating,ecoffet2019go}. Recent work in this family includes \citet{badia2020agent57}, \citet{bharadhwaj2020dynamics} and \citet{zhang2020automatic}.
Since the agent's entire policy changes with the goal, goal-conditioned exploration is somewhat similar to noise-conditioned \cite{plappert2017parameter,osband2017deep} and variational exploration algorithms (next paragraph), a key difference being that MEGA agents \textit{choose} their goal. 

\paragraph{Empowerment} Since the agent's off-policy, goal relabeling learning algorithm can be understood as minimizing the conditional entropy of (on-policy) achieved goals given some potential goal distribution $p_g$ (not necessarily the behavioural goal distribution $p_{bg}$), simultaneously choosing $p_{bg}$ to maximize entropy of historical achieved goals (the MEGA objective) results in an \textit{empowerment}-like objective:
$\max_{p_{bg}} H[p_{ag}] - H[\ag(\tau \given p_g) \given p_{g}] \approx \max_{p_{g}} I[p_{g}; \ag(\tau \given p_g)]$,
where equality is approximate because the first max is with respect to $p_{bg}$, and also because $H[p_{ag}]$ is historical, rather than on-policy. 

Empowerment \cite{klyubin2005empowerment, salge2014empowerment,mohamed2015variational} has gained traction in recent years as an intrinsic, unsupervised objective due to its intuitive interpretation and empirical success \cite{eysenbach2018diversity,hansen2019fast}. 
We can precisely define empowerment in the multi-goal case as the \textit{channel capacity} between goals and achieved goals \cite{cover2012elements}:
\begin{equation}\label{eq_empowerment}\small
	\mathcal{E}(s_0) = \max_{p_g} \mathbb{E}_{p(\tau  | g, s_0)p_g(g)} I[p_g; \ag(\tau \given p_g)],
\end{equation}
where $s_0$ represents the initial state distribution. 
To see the intuitive appeal of this objective, we reiterate the common argument and write: $I[p_g; \ag(\tau \given p_g)] = H[p_g] - H[p_g \given \ag(\tau\given p_g)],$
where $H$ is entropy. This now has an intuitive interpretation: letting $H[p_g]$ stand for the size of the goal set, and $H[p_g \given \ag(\tau\given p_g)]$ for the uncertainty of goal achievement, maximizing empowerment roughly amounts to maximizing the \textit{size of the achievable goal set}.

The common approach to maximizing empowerment has been to either fix or parameterize the distribution $p_g$ and maximize the objective $I[p_g; \ag(\tau\given p_g)]$ \textit{on-policy} \cite{gregor2016variational,warde-farley2018unsupervised,pong2019skew}. We can think of this as  approximating (\ref{eq_empowerment}) using the behavioural goal distribution $p_{bg} \approx \arg\max_{p_g} I[p_g; \ag(s_T\given p_g)]$. A key insight behind our work is that there is no reason for an off-policy agent to constrain itself to pursuing goals from the distribution it is trying to optimize. Instead, we argue that for off-policy agents seeking to optimize (\ref{eq_empowerment}), the role of the behavioural goal distribution $p_{bg}$ should be to produce useful empirical data for optimizing the true \textit{off-policy} empowerment (\ref{eq_empowerment}), where the maximum is taken over all possible $p_g$. Practically speaking, this means exploring to maximize entropy of the historical achieved goal distribution (i.e,. the MEGA objective), and letting our off-policy, goal relabeling algorithm minimize the conditional entropy term. Future work should investigate whether the off-policy gain of MEGA over the on-policy Diverse sampling can be transferred to general empowerment maximizing algorithms. 

\section{Limitations and Future Work}

The present work has several limitations that should be addressed by future work. First, our experiments focus on environments with predefined, semantically meaningful, and well-behaved goal spaces. In the general case, an agent will have to learn its own goal space \cite{warde-farley2018unsupervised,pong2019skew} and it will be interesting to see whether MEGA exploration extends well to latent spaces. A foreseeable problem, which we did not experience, is that differential entropy is sensitive to reparameterizations of the feature space; this implies that either (1) a MEGA agent's goal space needs to be, to a degree, ``well-behaved'', or (2) the MEGA objective needs to be recast or extended so as to be robust to parameterization. We hypothesize that a major reason for MEGA's success is that the goal spaces in our chosen tasks are semantically meaningfully and directly relevant to the tasks being solved; an interesting direction for future research involves the automatic discovery of such low-dimensional abstractions  \cite{lee2020weakly}. A second limitation of our work is the approach to achievability, which is required in order for our minimum density heuristic to be sensible. Presently, we rely on a combination of buffer-based generation, a cutoff mechanism that eliminates goals with low Q-values, and the ability of our off-policy learning algorithm to generalize. But even so, our \env{FetchStack2} results show that the MEGA agent's performance begins to diverge after about 5 million steps. This is because the table (on which the blocks are being stacked) is not enclosed, and the agent begins to pursue difficult to achieve goals that are off the table. Future work should explore better ways to measure off-policy achievability \cite{thomas2015high}. Finally, the behavior of MEGA on \env{FetchStack2} suggests that an unconstrained, intrinsically motivated agent may start to set unsafe goals, which has implications for safety \cite{garcia2015comprehensive}.

\section{Conclusion} \label{section_conclusion}

This paper proposes to address the long-horizon, sparse reward problem in multi-goal RL by having the agent maximize the entropy of the historical achieved goal distribution. We do this by setting intrinsic goals in sparsely explored areas of the goal space, which focuses exploration on the frontier of the achievable goal set. This strategy obtains results that are more than 10 times more sample efficient than prior approaches in four long-horizon multi-goal tasks. 

\vfill
\section*{Acknowledgments}

We thank Sheng Jia, Michael R. Zhang and the anonymous reviewers for their helpful comments. Resources used in preparing this research were provided, in part, by the Province of Ontario, the Government of Canada through CIFAR, and companies sponsoring the Vector Institute (\url{https://vectorinstitute.ai/#partners}).


{
\bibliography{refs}
\bibliographystyle{icml2020_style/icml2020}
}


\newif\iflong
\longtrue 

\vfill
\newpage

\appendix

\onecolumn

\iflong%
\else
\section*{Abridged Appendix}
\fi

\section{Comparison of Functions Used by Multi-Goal Agents}
\label{appendix_agentcomparison}

\newcolumntype{L}[1]{>{\raggedright\let\newline\\\arraybackslash\hspace{0pt}}m{#1}}
\renewcommand\tabularxcolumn[1]{m{#1}}
\newcolumntype{C}{>{\centering\arraybackslash}X}
\begin{table}[h]\scriptsize
\begin{tabularx}{\textwidth}{L{1.15in}|Xm{0.01in}m{1.6in}m{0.01in}m{1.35in}}
 \toprule
 \ &  \textsc{Select} & &\textsc{Reward} & &\textsc{Relabel} \\
 \toprule 
   \makecell[l]{HER\\ \cite{andrychowicz2017hindsight} }&     Samples from environment  & &    Sparse environment reward   & &   \strat{future}     \\\midrule
   \makecell[l]{SAGG-RIAC\\ \cite{baranes2013active} }&  
   Chooses goals in areas where absolute learning progress is greatest &&
   Uses ``competence'', defined as the normalized negative Euclidean distance between final achieved goal and goal. 
   &  &   N/A    \\\midrule
   \makecell[l]{Goal GAN\\ \cite{florensa2018automatic} }& 
   Samples from a GAN trained to generate goals of intermediate difficulty && 
   Sparse environment reward   & &    
   N/A       \\\midrule
   \makecell[l]{RIG \\ \cite{nair2018visual} }& 
   Samples from the prior of a generative model (VAE) fitted to past achieved goals&&  
   $r(s,g) = - \norm{e(s) - e(g)}$, Negative Euclidean distance in latent space & & 
   50\% \strat{future}, 50\% samples from generative model \\\midrule
   \makecell[l]{DISCERN\\ \cite{warde-farley2018unsupervised} }& 
   Samples uniformly from a diversified buffer of past achieved goals &&      
   $r_T = \max(0, l_g)$, where $l_g$ is dot product between L2-normalized embeddings of the state and goal, and $r_t = 0$ otherwise for $t=0,\ldots,T-1$ &&
   Samples \mbox{$g'\!\in\!\{g'_{T-H},\dots,g'_{T}\}$} and sets $r_T = 1$ \\\midrule
   \makecell[l]{CURIOUS\\ \cite{colas2018curious} }& 
   Samples from one of several environments in which absolute learning progress is greatest& & 
   Sparse environment reward depending on the current environment (module/task)   &  &
   \strat{future}   \\\midrule
   \makecell[l]{CHER\\ \cite{fang2019curriculum} }&  Samples from environment  & & 
   Sparse environment reward & &
   Select based on sum of diversity score of selected goals and proximity score to desired goals\\\midrule
   \makecell[l]{Skew-Fit\\ \cite{pong2019skew} }&  Samples from a generative model that is skewed to be approximately uniform over past achieved goals & &  $r(s,g) = - \norm{e(s) - e(g)}$, Negative Euclidean distance in latent space  &  & 50\% \strat{future}, 50\% samples from generative model        \\\midrule
   \makecell[l]{DDLUS\\ \cite{hartikainen2020dynamical} }&   
   Selects maximum distance goals according to a learned distance function & &  
   $r(s,g) = -d^{\pi}(s, g)$, \mbox{Negative} expected number of time steps for a policy $\pi$ reach goal $g$ from state $s$ 
   & &   N/A     \\\midrule
   \makecell[l]{MEGA\\ (Ours, 2020) }&    
   Selects low density goals according to a learned density model  & &   
   Sparse environment reward      &   & 
   \strat{rfaab} (Appendix \ref{appendix_implementation})   \\
\bottomrule
\end{tabularx}
    \caption{High level summary of \textsc{Select}, \textsc{Reward}, \textsc{Relabel} functions used by various multi-goal algorithms for Algorithm \ref{alg_goal_seeking}.}\label{tab:my_label}
\end{table}

\iflong%
\else
\begin{multicols}{2}
\fi

\section{Worked Propositions}

\iflong
\setcounter{proposition}{0}
\label{appendix_props}

\begin{proposition}[Discrete Entropy Gain]
Given buffer $\mathcal{B}$ with $\eta\!=\!\frac{1}{|\mathcal{B}|}$, maximizing expected next step entropy is equivalent to maximizing expected point-wise entropy gain $\Delta H (g')$:
\begin{equation}\label{eq_prop_entropy_gain}
\begin{split}\small
 \hat g^* &=\arg\max_{\hat{g} \in \mathcal{B}} \mathbb{E}_{g' \sim q(g' \given \hat g)} H[p_{ag \given g'}] \\
 &= \arg\max_{\hat{g}\in \mathcal{B}} \mathbb{E}_{g' \sim q(g' \given \hat g)} \Delta H (g'),
\end{split}
\end{equation}
where $\Delta H (g') = p_{ag}(g') \log p_{ag}(g')\ - (p_{ag}(g') + \eta) \log (p_{ag}(g') + \eta)$.
\end{proposition}

\begin{proof}
Expanding the expression for maximizing the expected next step entropy:
\begin{align} \label{eqn:entropy_gain_obj}
    \hat{g}^* &= \argmax_{\hat{g} \in \mathcal{B}} \E_{g' \sim q(g'|\hat{g})} H[p_{ag \given g'}] \\
    &= \argmax_{\hat{g}\in \mathcal{B}} \sum_{g'} q(g'|\hat{g}) \sum_{g} -p_{g'}(g) \log p_{g'}(g)
\end{align}

We write the new empirical achieved goal distribution $p_{ag \given g'}(g)$ after observing achieved goal $g'$ in terms of $ \eta\!=\!\frac{1}{|\mathcal{B}|}$ and the original achieved goal distribution $p_{ag}(g)$:
\begin{align}
    p_{ag \given g'}(g) = \frac{p_{ag}(g) + \eta \mathds{1}[g=g']}{1 + \eta}
\end{align}
Substituting the expression of $p_{ag \given g'}(g)$ into above, we can ignore several constants when taking the $\argmax$ operation:
\begin{align}
    \hat{g}^*  &= \argmax_{\hat{g}\in \mathcal{B}} \sum_{g'} q(g'|\hat{g}) \sum_{g} -\frac{p_{ag}(g) + \eta \mathds{1}[g=g']}{1 + \eta} \log \frac{p_{ag}(g) + \eta \mathds{1}[g=g']}{1 + \eta} \\
    &= \argmax_{\hat{g}\in \mathcal{B}} \sum_{g'} q(g'|\hat{g}) \sum_{g} -(p_{ag}(g) + \eta \mathds{1}[g=g']) \log (p_{ag}(g) + \eta \mathds{1}[g=g'])
\end{align}

We can break down the summation for the new entropy $H[p_{ag \given g'}] = H[p_{ag}] + \Delta H(g')$ as the difference between the original entropy $H[p_{ag}]$ and the difference term $\Delta H(g')$. Then we can discard the original entropy term which is constant with respect to the sampled achieved goal $g'$:
\begin{align}
    \hat{g}^* &= \argmax_{\hat{g}\in \mathcal{B}} \sum_{g'} q(g'|\hat{g}) [H[p_{ag}] + \Delta H(g') ] \\
    &= \argmax_{\hat{g}\in \mathcal{B}} \sum_{g'} q(g'|\hat{g}) \big[ \big(-\sum_{g} p_{ag}(g) \log p_{ag}(g) \big) + \big(p_{ag}(g') \log p_{ag}(g') - (p_{ag}(g') + \eta) \log (p_{ag}(g') + \eta) \big) \big] \\
    &= \argmax_{\hat{g}\in \mathcal{B}} \sum_{g'} q(g'|\hat{g}) \big[p_{ag}(g') \log p_{ag}(g') - (p_{ag}(g') + \eta) \log (p_{ag}(g') + \eta) \big] \\
    &= \argmax_{\hat{g}\in \mathcal{B}} \E_{g' \sim q(g'|\hat{g})} \big[p_{ag}(g') \log p_{ag}(g') - (p_{ag}(g') + \eta) \log (p_{ag}(g') + \eta) \big] \label{eqn:entropy_gain_obj_2}
\end{align}

Therefore, we have reduced the complexity for computing $\hat{g}^*$ with Equation (\ref{eqn:entropy_gain_obj_2}) to $O(d^2)$ from $O(d^3)$ in Equation (\ref{eqn:entropy_gain_obj}), where $d$ denotes the size of the support. 

\end{proof}

\begin{proposition}[Discrete Entropy Gradient]\label{prop_discrete_entropy_gradient}
\begin{equation}\small
\begin{split}
    \lim_{\eta \to 0} \hat g^* &= \arg\max_{\hat{g}\in \mathcal{B}}\langle \nabla_{p_{ag}} H[p_{ag}], q(g' \given \hat g) - p_{ag} \rangle\\
    &= \arg\max_{\hat{g}\in \mathcal{B}} \DKL{q(g' \given \hat g)}{p_{ag}} + H[q(g' \given \hat g)]
\end{split}%
\end{equation}%
\end{proposition}%

\begin{proof}
Putting $p = p_{ag}(g')$ and dividing by $\eta$, we have: 
\begin{equation}\small
\begin{split}
    \lim_{\eta \to 0} \frac{\Delta H(g')}{\eta} 
    &= \lim_{\eta \to 0}\frac{p}{\alpha} \log p - \frac{1}{\alpha}(p + \eta) \log (p + \eta)\\
    &= \lim_{\eta \to 0}-\log\left(\frac{p+\eta}{p}\right)^{p/\eta} - \log (p + \eta)\\
    &= -\log e - \log p\\
    &= - 1 - \log p.
\end{split}
\end{equation}
This is the same as $\nabla_{p_{ag}} H[p_{ag}]$, so that: $$\lim_{\alpha \to 0} \hat g^* =  \arg\max_{\hat{g}\in \mathcal{B}} \mathbb{E}_{g' \sim q(g' \given \hat g)} \nabla_{p_{ag}} H[p_{ag}](g') = \arg\max_{\hat{g}\in \mathcal{B}} \langle \nabla_{p_{ag}} H[p_{ag}], q(g' \given \hat g) \rangle.$$
Then, to get the first equality, we subtract $p_{ag}$ from $q(g' \given \hat g)$ inside the inner product since it does not depend on $\hat g$, and results in a directional derivative that respects the constraint $\int\!p = 1$. The second equality follows easily after substituting $\nabla_{p_{ag}} H[p_{ag}] = - 1 - \log p$ into $\langle \nabla_{p_{ag}} H[p_{ag}], q(g' \given \hat g) - p_{ag}\rangle$.\end{proof}

Let us now assume that the empirical $p_{ag}$ is used to induce (abusing notation) a density $p_{ag}$ with full support. This can be done by assuming that achieved goal observations are noisy (note: this is not the same thing as relaxing the Markov state assumption) and using $p_{ag}$ to represent our posterior of the goals that were actually achieved. Then, Proposition \ref{prop_discrete_entropy_gradient} extends to the continuous case by taking the functional derivative of differential entropy with respect to the variation $\eta(g) = q(g' \given \hat g)(g) - p_{ag}(g)$. We consider only the univariate case $G = \R$ below. 

\newtheorem{innercustomprop}{Proposition}
\newenvironment{customprop}[1]
  {\renewcommand\theinnercustomprop{#1}\innercustomprop}
  {\endinnercustomprop}
  
\begin{customprop}{2$^*$}[Differential Entropy Gradient]
\begin{equation}
    \delta H(p_{ag}, \eta) \triangleq \lim_{\epsilon \to 0}\frac{H(p_{ag}(g) + \epsilon \eta(g)) - H(p_{ag}(g))}{\epsilon}= \DKL{q(g' \given \hat g)}{p_{ag}} + H[q(g' \given \hat g)] - H[p_{ag}]
\end{equation}
\end{customprop}
\begin{proof}
Since the derivative of $f(x) = x \log x$ is $1 + \log x$, the variation $\delta H(p_{ag}, \eta)$ with respect to $\eta$ is:
\begin{equation}\label{eq_first_variation}
    \begin{split}
    \delta H(p_{ag}, \eta) &= -\lim_{\epsilon \to 0}\frac{1}{\epsilon}\int_{-\infty}^\infty (p_{ag} + \epsilon\eta) \log (p_{ag} + \epsilon\eta) - (p_{ag} \log p_{ag}) dx\\
    &= \lim_{\epsilon \to 0}\frac{1}{\epsilon}\int_{-\infty}^\infty (-1 - \log p_{ag})\epsilon\eta + O((\epsilon\eta)^2) dx\\
    &= \lim_{\epsilon \to 0}\frac{1}{\epsilon}\int_{-\infty}^\infty (-1 - \log p_{ag})\epsilon\eta dx  + \lim_{\epsilon \to 0}\frac{1}{\epsilon} \int_{-\infty}^\infty O((\epsilon\eta)^2) dx\\
    &= \int_{-\infty}^\infty (-1 - \log p_{ag})(q(g' \given \hat g) - p_{ag})  dx \\
    &= -\mathbb{E}_{q(g' \given \hat g)}\log p_{ag}(x) + \mathbb{E}_{p_{ag}}\log p_{ag}(x)\\[6pt]
    &= \DKL{q(g' \given \hat g)}{p_{ag}} + H[q(g' \given \hat g)] - H[p_{ag}]
    \end{split}
\end{equation}
where the second equality uses Taylor's theorem, and the third equality is justified when the limits exist. As in Proposition \ref{prop_discrete_entropy_gradient}, this functional derivative is maximized by choosing $q(g' \given \hat g)$ to maximize $\DKL{q(g' \given \hat g)}{p_{ag}} + H[q(g' \given \hat g)]$. 
\end{proof}

\begin{proposition}\label{prop_det_cond_approx}
If $q(g'| \hat g) = \mathds{1}[g' = \hat g]$, the discrete entropy gradient objective simplifies to a minimum density objective:
\begin{equation}\small
\begin{split}
    \hat g^* 
    &= \arg\max_{\hat{g}\in \mathcal{B}} -\log [p_{ag}(\hat g)] \\
    &= \arg\min_{\hat{g}\in \mathcal{B}} p_{ag}(\hat g)
\end{split}%
\end{equation}%
\end{proposition}%

\begin{proof}
We substitute the the case where $q(g'| \hat g) = \mathds{1}[g' = \hat g]$ into the discrete entropy objective and simplify:
\begin{align}
    \lim_{\eta \to 0} \hat g^* &= \arg\max_{\hat{g}\in \mathcal{B}} \DKL{q(g' \given \hat g)}{p_{ag}} + H[q(g' \given \hat g)] \\
    &= \arg\max_{\hat{g}\in \mathcal{B}} \DKL{\mathds{1}[g' = \hat g]}{p_{ag}} + H[\mathds{1}[g' = \hat g]] \\
    &= \arg\max_{\hat{g}\in \mathcal{B}} \big( \sum_{g'} \mathds{1}[g' = \hat g](\log \mathds{1}[g' = \hat g] - \log p_{ag}(g')) \big) + 0 \\
    &= \arg\max_{\hat{g}\in \mathcal{B}} \log 1 - \log p_{ag}(\hat g) \\
    &= \arg\max_{\hat{g}\in \mathcal{B}} - \log p_{ag}(\hat g) \\
    &= \arg\min_{\hat{g}\in \mathcal{B}} p_{ag}(\hat g)
\end{align}

In particular, we eliminate the entropy term $H[\mathds{1}[g' = \hat g]] = 0$, simplify the sum to only consider $g' = \hat g$ term, and note that maximizing the negative log probability is equivalent to finding the minimum probability. 
\end{proof}

\else
Omitted in abridged version.
\fi

\section{Implementation Details}\label{appendix_implementation}

\paragraph{Code} 

Code to reproduce all experiments is available at \url{https://github.com/spitis/mrl} \cite{mrl}.

\paragraph{Base Implementation} 

We use a single, standard DDPG agent \cite{lillicrap2015continuous} that acts in multiple parallel environments (using Baseline's VecEnv wrapper \cite{baselines}). Our agent keeps a single replay buffer and copy of its parameters and training is parallelized using a GPU. We utilize many of the same tricks as \citet{plappert2018multi}, including clipping raw observations to [-200, 200], normalizing clipped observations, clipping normalized observations to [-5, 5], and clipping the Bellman targets to [$-\frac{1}{1-\gamma}$, 0]. Our agent uses independently parameterized, layer-normalized \cite{ba2016layer} actor and critic, each with 3 layers of 512 neurons with GeLU activations \cite{hendrycks2016gaussian}. We apply gradient value clipping of 5, and apply action l2 regularization with coefficient 1e-1 to the unscaled output of the actor (i.e., so that each action dimension lies in [-1., 1.]). No weight decay is applied. We apply action noise of 0.1 to the actor at all exploration steps, and also apply epsilon random exploration of 0.1. Each time the goal is achieved during an exploratory episode, we increase the amount of epsilon random exploration by 0.1 (see Go Exploration below). We use a discount factor ($\gamma$) of 0.98 in \env{PointMaze} (50 steps / episode), \env{FetchPickAndPlace} (50 steps / episode), and \env{FetchStack2} (50 steps / episode), and a discount factor of 0.99 in the longer horizon \env{Antmaze} (500 steps / episode). 

\paragraph{Optimization}

We use Adam Optimizer \cite{kingma2014adam} with a learning rate of 1e-3 for both actor and critic, and update the target networks every 40 training steps with a Polyak averaging coefficient of 0.05. We vary the frequency of training depending on the environment, which can stabilize training: we optimize every step in \env{PointMaze}, every two steps in \env{Antmaze}, every four steps in \env{FetchPickAndPlace}, and every ten steps in \env{FetchStack2}. Since optimization occurs every $n$ environment steps, regardless of the number of environments (typically between 3 and 10), using a different number of parallel environments has neglible impact on performance, although we kept this number the same between different baselines. Optimization steps use a batch size of 2000, which is sampled uniformly from the buffer (no priorization). There is an initial ``policy warm-up'' period of 5000 steps, during which the agent acts randomly. Our replay buffer is of infinite length (this is obviously inefficient, and one should considering prioritized sampling \cite{schaul2015prioritized} and diverse pruning of the buffer \cite{abels2018dynamic} for tasks with longer training horizons). 

\paragraph{Goal Relabeling} \label{appendix_rfaab}

We generalize the \texttt{future} strategy by additionally relabeling transitions with goals randomly sampled (uniformly) from buffers of \strat{actual} goals (i.e., a buffer of the past desired goals communicated to the agent at the start of each episode), past \strat{achieved} goals, and behavioral goals (i.e., goals that agent pursues during training, whether intrinsic or extrinsic). We call this the \strat{rfaab} strategy, which stands for Real (do not relabel), Future, Actual, Achieved, and Behavioral. Intuitively, relabeling transitions with goals outside the current trajectory allows the agent to generalize across trajectories. Relabeling with actual goals focuses optimization on goals from the desired goal distribution.  Relabeling with achieved goals could potentially maintain performance with respect to past achieved goals. Relabeling with behavioral goals focuses optimization on past intrinsic goals, potentially allowing the agent to master them faster. 

All relabeling is done online using an efficient, parallelized implementation. The \strat{rfaab} strategy requires, as hyperparameters, relative ratios of each kind of experience, as it will appear in the minibatch. Thus, \strat{rfaab\_1\_4\_3\_1\_1} keeps 10\% real experiences, and relabels approximately 40\% with \strat{future}, 30\% with \strat{actual}, 10\% with \strat{achieved}, and 10\% with \strat{behavioral}. 
\iflong
The precise number of relabeled experiences in each minibatch is sampled from the appropriate multinomial distribution (i.e., with $k=5$, $n=$ batch size, and the \strat{rfaab} ratios). Note that \strat{rfaab} is a strict generalization of \strat{future}, and that the \strat{future\_4} strategy used by \citet{andrychowicz2017hindsight} and \citet{plappert2018multi} is the same as \strat{rfaab\_1\_4\_0\_0\_0}. Because of this, we do not do a thorough ablation of the benefit over future, simply noting that others have found \strat{achieved}  (e.g., \citet{nair2018visual}) and \strat{actual} (e.g., \citet{pitisprotoge}) sampling to be effective; rather, we simply conducted a random search over the \strat{rfaab} hyperparameters that included pure \strat{future} settings (see Hyperparameter Selection below). We found in preliminary experiments that using a pure \strat{future} warmup (i.e., 100\% of experiences relabeled using future achieved goals from the same trajectory) helped with early training, and relabeled goals using this strategy for the first 25,000 environment steps before switching to \strat{rfaab}.
\else
We use \strat{rfaab\_1\_4\_3\_1\_1} in \env{PointMaze} and \env{Antmaze} and \strat{rfaab\_1\_5\_2\_1\_1}. in \env{Fetch} environments. Some details omitted in abridged.
\fi

\paragraph{Density Modeling}

\iflong
\begin{wrapfigure}{r}{0.4\textwidth}
\vspace{-0.2in}
  \begin{center}
    \includegraphics[width=0.38\textwidth]{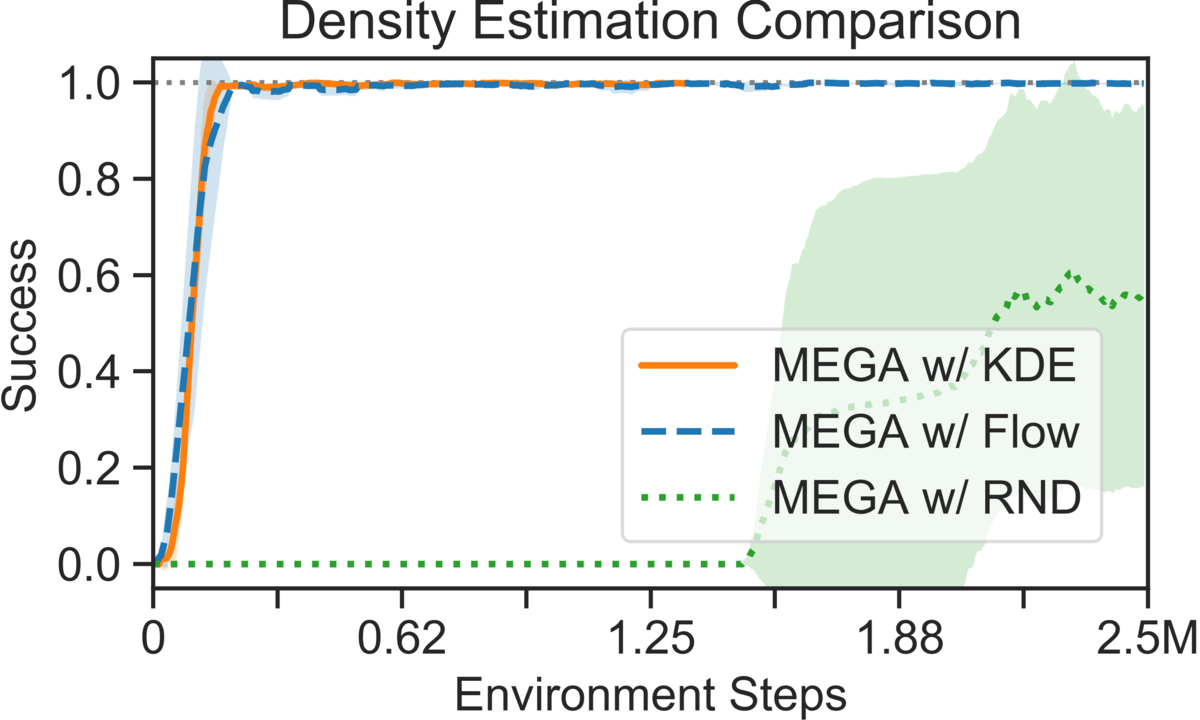}
  \end{center}
  \vspace{-0.2in}
  \caption{\small Density modeling on \env{Pointmaze}.}\label{fig_density_estimation}
    \vspace{-0.2in}
\end{wrapfigure}
\fi 

We considered three approaches to density modeling: a kernel density estimator (KDE) \cite{rosenblatt1956remarks}, a normalizing flow (Flow) \cite{papamakarios2019normalizing} based on RealNVP \cite{dinh2016density}, and a random network distillation (RND) approach \cite{burda2018exploration}. Based on the resulting performances and relatively complexity, we chose to use KDE throughout our experiments. 

\iflong
The KDE approach fits the KDE density model from Scikit-learn \cite{pedregosa2011scikit} to normalized samples from the achieved goal distribution (and additionally, in case of OMEGA, the desired goal distribution). Since samples are normalized, the default bandwidth (0.1) and kernel (Gaussian) are good choices, although we ran a random search (see Hyperparameter Selection) to confirm. The model is computationally inexpensive, and so we refit the model on every optimization step using 10,000 normalized samples, sampled uniformly from the buffer.

The Flow approach optimizes a RealNVP model online, by training on a mini-batch of 1000 samples from the buffer every 2 optimization steps. It uses 3 pairs of alternating affine coupling layers, with the nonlinear functions modeled by 2 layer fully connected neural networks of 256 leaky ReLU neurons. It is optimized using Adam optimizer with a learning rate of 1e-3.

The RND approach does not train a true density estimator, and is only intended as an approximation to the minimum density (by evaluating relative density according to the error in predicting a randomly initialized network). Both our random network and learning network were modeled using 2 layer fully connected neural networks of 256 GeLU neurons. The learning network was optimized online, every step using a batch size of 256 and stochastic gradient descent, with learning rate of 0.1 and weight decay of 1e-5. 
\fi

We tested each approach in \env{PointMaze} only
\iflong
---the results are shown in Figure \ref{fig_density_estimation}. 
\fi
Both the KDE and Flow models obtain similar performance, whereas the RND model makes very slow progress. Between KDE and Flow, we opted to use KDE throughout our experiments as it is fast, easy to understand and implement, and equally effective in the chosen goal spaces (maximum 6 dimensions). It is possible that a Flow (or VAE-like model \cite{nair2018visual}) would be necessary in a higher dimensional space, and we expect that RND will work better in high dimensional spaces as well.  

\paragraph{Cutoff mechanism and Go Exploration}

\iflong

As noted in the main text, the idea behind our minimum density heuristic is to quickly reachieve a past achieved, low density goal, and explore around (and hopefully beyond) it \cite{ecoffet2019go}. If the agent can do this, its conditional achieved goal distribution $q(g' \given \hat g)$ will optimize the MEGA objective well (see Propositions). This requires two things: first, that behavioral goals be achievable, and second, that the agent explores around them, and doesn't simply remain in place once they are achieved. 

To ensure that goal are achievable, we use a simple cutoff mechanism that crudely prevents the agent from attempting unobtainable goals, as determined by its critic. During \textsc{Select} function (Algorithm \ref{alg_mega_select}, reproduced below), the agent ``eliminates unachieved candidates'' using this mechanism. This amounts to rejecting any goal candidates whose Q-values, according to the critic, are below the current cutoff. The current cutoff is initialized at -3., and decreased by 1 every time the agent has an ``intrinsic success percentage'' of more than 70\% over the last 10 training episodes, but never below the minimum of the lowest Q-value in sampled goal candidates. It is increased by 1 every time the agent's intrinsic success percentage is less than 30\% over the last 10 training episodes. An agent is intrinsically successful when it achieves its behavioral goal at least one time, on any time step, during training (using its exploratory policy). We can see the intrinsic success for MEGA, OMEGA, and the various baselines in Figure \ref{fig:intrinsic_successes}. Although we have found that this cutoff mechanism is not necessary and in most cases neither helps nor hurts performance, it helps considerably in certain circumstances (especially, e.g., if we add the environment's desired goal to the candidate goal set in the \textsc{MEGA\_Select} function of Algorithm \ref{alg_mega_select}). See ablations in Appendix \ref{apdx_ablation}. We applied this same cutoff mechanism to all baselines except the Goal Discriminator baseline, which has its own built-in approach to determining goal achievability. Note from Figure \ref{fig:intrinsic_successes}, however, that the cutoff was rarely utilized for non-MEGA algorithms, since they all quickly obtain high intrinsic success percentages (this is a direct consequence of their lower achieved goal entropy: since they explore less of the goal space in an equal amout of time, they have relatively more time to optimize their known goal spaces---e.g., Achieved sampling maintains close to 100\% intrinsic success even in \env{Stack2}). Note that due to the cutoff mechanism, MEGA's intrinsic success hovers around 50\% in the open-ended \env{Fetch} environments, where the block is often hit off the table and onto the floor and so achieved very difficult to re-achieve goals.

\begin{figure}[!t]
    \begin{subfigure}[b]{0.24\textwidth}
         \centering
         \includegraphics[width=\textwidth]{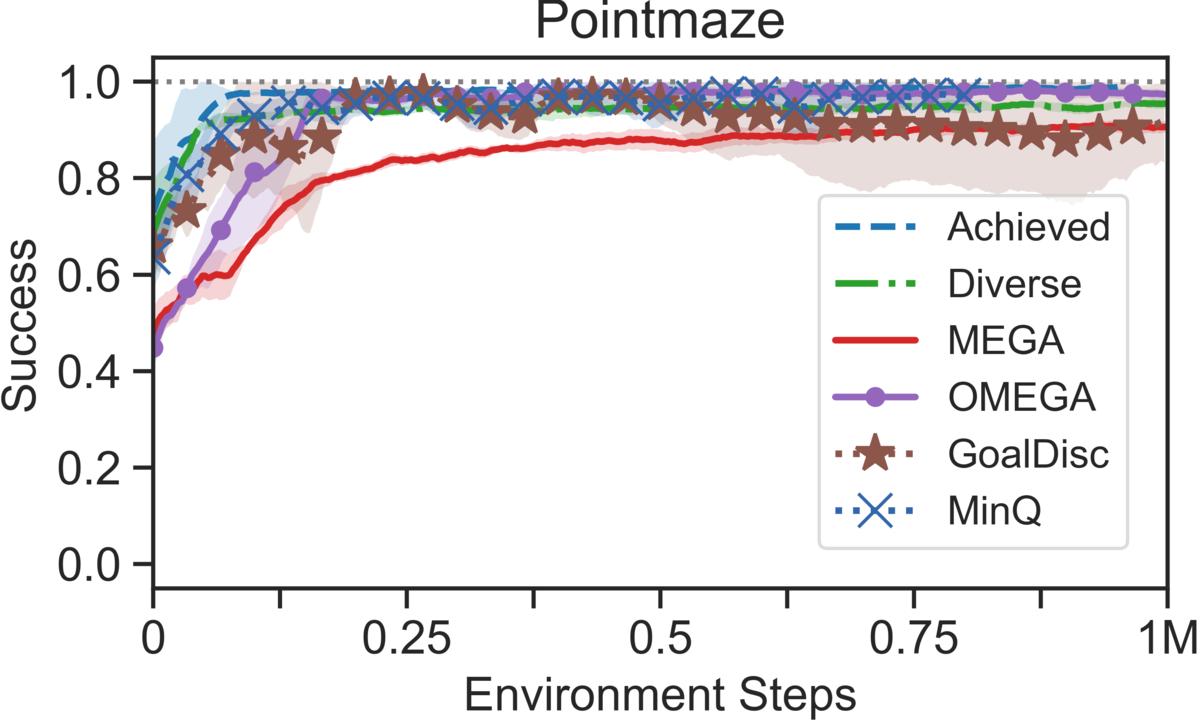}
         \label{fig:}
     \end{subfigure}
     \hfill
     \begin{subfigure}[b]{0.24\textwidth}
         \centering
         \includegraphics[width=\textwidth]{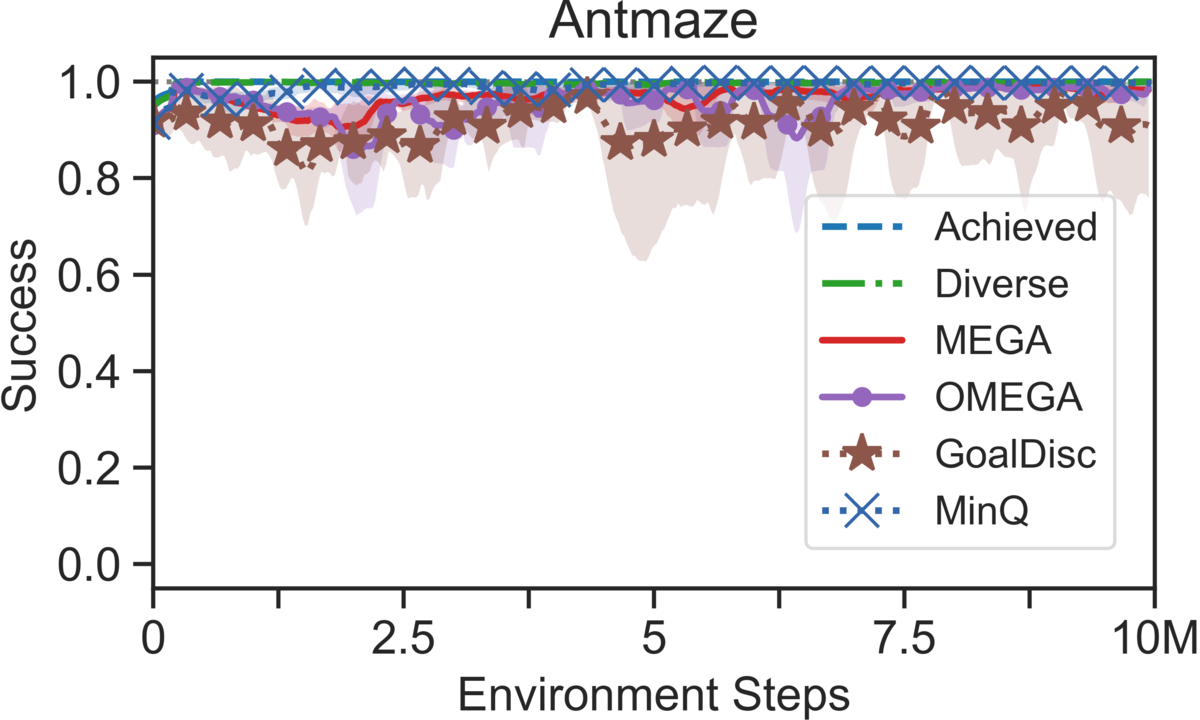}
         \label{fig:}
     \end{subfigure}
     \hfill
     \begin{subfigure}[b]{0.24\textwidth}
         \centering
         \includegraphics[width=\textwidth]{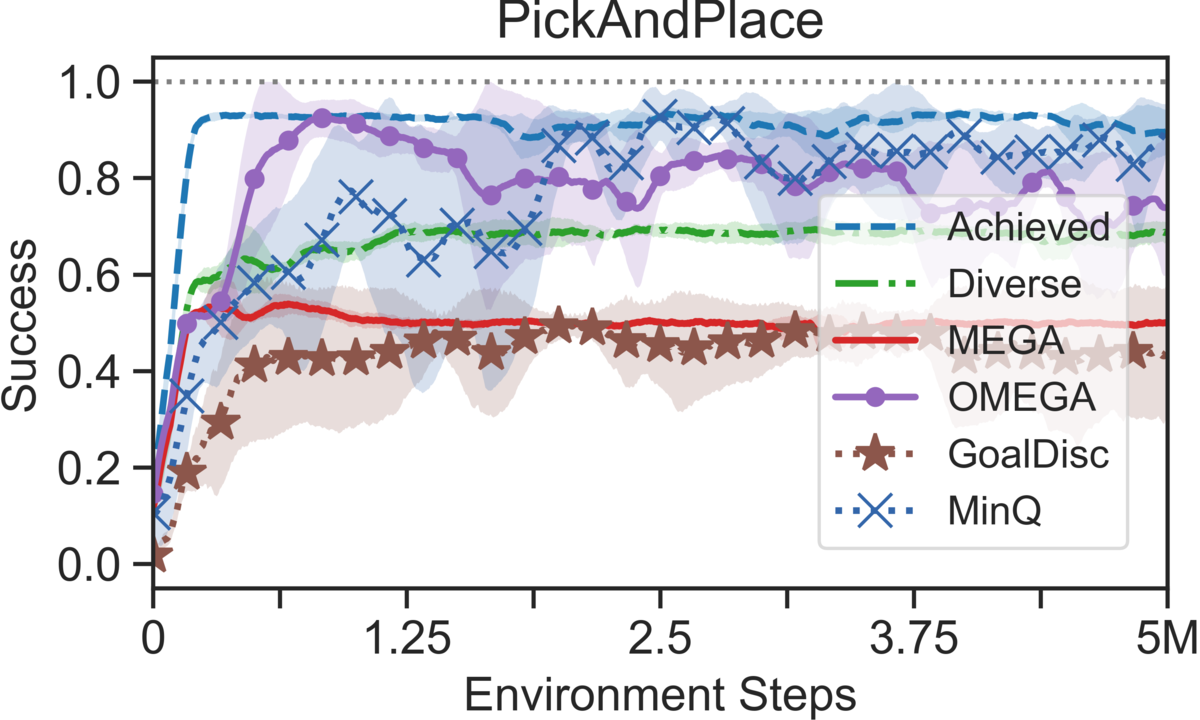}
         \label{fig:}
     \end{subfigure}
     \hfill
     \begin{subfigure}[b]{0.24\textwidth}
         \centering
         \includegraphics[width=\textwidth]{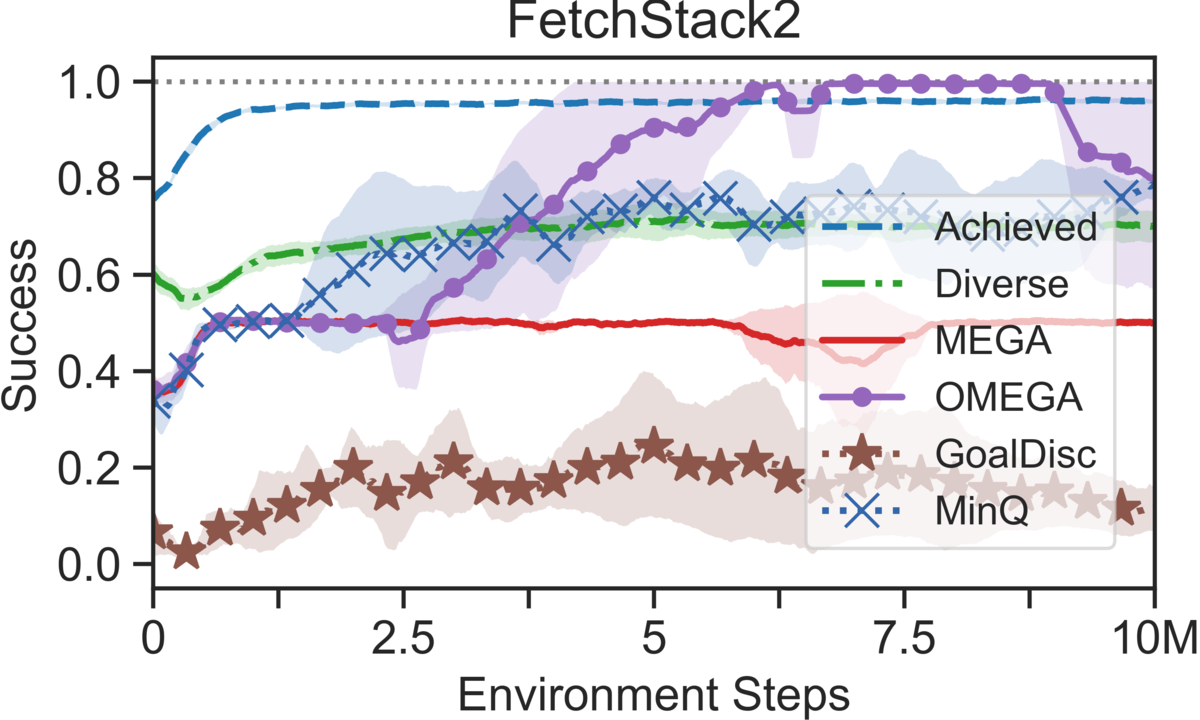}
         \label{fig:}
     \end{subfigure}
     \vspace{-0.3in}
     \caption{Intrinsic success in various environments, which is the proportion of \textit{training} episodes that are successful on \textit{first visit basis}---that is, training episodes in which the agent achieves its behavioral goal at least one time, on any timestep, using its exploratory policy.}
    \label{fig:intrinsic_successes}
\end{figure}

To encourage the agent to explore around the behavioral goal, we increase the agent's exploratory behaviors every time the behavioral goal is reachieved in any given episode. We refer to this as ``Go Exploration'' after \citet{ecoffet2019go}, who used a similar approach to reset the environment to a frontier state, and explored around that state. We use a very simple exploration bonus, which increases the agent's epsilon exploration by a fixed percentage. We use 10\% (see next paragraph), which means that an agent which accomplishes the achieved goal 10 times in an episode will be exploring purely at random. Intuitively, this corresponds to ``filling in'' sparse regions of the achieved goal space (see Figure \ref{fig_filling_in}). Likely there are more intelligent ways to do Go Exploration, but we leave this for future work. Note that all baselines benefited from this feature. 

\begin{figure}[!h]
    \begin{subfigure}[b]{0.24\textwidth}
         \centering
         \includegraphics[width=\textwidth]{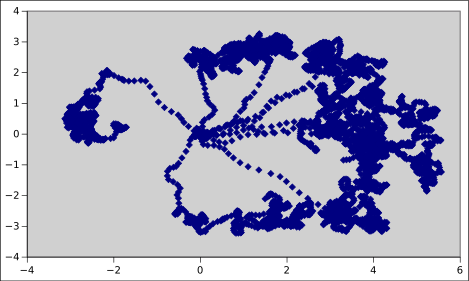}
         \label{fig:}
     \end{subfigure}
     \hfill
     \begin{subfigure}[b]{0.24\textwidth}
         \centering
         \includegraphics[width=\textwidth]{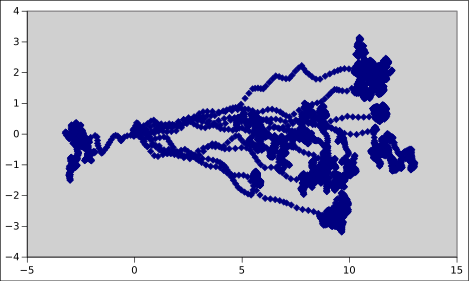}
         \label{fig:}
     \end{subfigure}
     \hfill
     \begin{subfigure}[b]{0.24\textwidth}
         \centering
         \includegraphics[width=\textwidth]{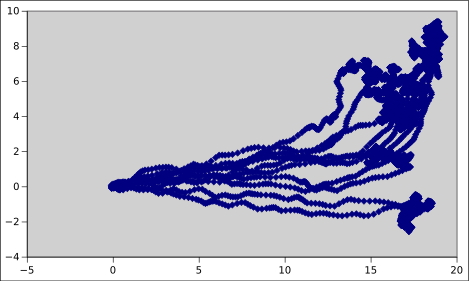}
         \label{fig:}
     \end{subfigure}
     \hfill
     \begin{subfigure}[b]{0.24\textwidth}
         \centering
         \includegraphics[width=\textwidth]{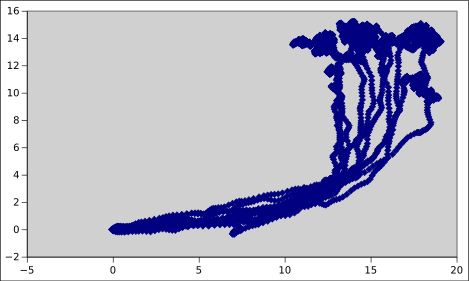}
         \label{fig:}
     \end{subfigure}
     \vspace{-0.2in}
     \caption{Plots of the last 10 trajectories leading up to 250K, 500K, 750K and 1M environment steps (from left to right) in \env{Antmaze}. Note that the scales on each plot are different. We see that the agent approximately travels directly to the achieved goal (with some randomness due to exploration noise), at which point it starts to explore rather randomly, forming a flowery pattern and ``filling in'' the low density region.}
    \label{fig_filling_in}
\end{figure}

\else
Omitted in abridged.
\fi

\newcommand{\taub}{\bs{\tau}}
\paragraph{Maximum Entropy-based Prioritization (MEP)} We implemented MEP \citep{zhao2019maximum} in our code base to compare the effects of entropy-based prioritization during the \textsc{OPTIMIZE} step while inheriting the rest of techniques described earlier. We based our implementation of MEP on the official repository\footnote{\url{https://github.com/ruizhaogit/mep}}. We represented the trajectory as a concatenation of the achieved goal vectors over the episode with $T$ timesteps, $\tau = [g'_0; g'_1; \dots ;g'_T]$.  To model the probability density of the trajectory of achieved goals, we used a Mixture of Gaussians model: 
\begin{equation}
p(\tau \mid \phi)= \frac{1}{Z} \sum_{i=k}^K c_k \mathcal{N}(\tau | \boldsymbol{\mu}_k, \boldsymbol{\Sigma}_k),
\label{eq:mep_mog}
\end{equation}
where we denote each Gaussian as $\mathcal{N}(\tau | \boldsymbol{\mu}_k, \boldsymbol{\Sigma}_k)$, consisting of its mean $\boldsymbol{\mu}_k$ and covariance $\boldsymbol{\Sigma}_k$. $c_k$ refers to the mixing coefficients and $Z$ to the partition function. $\phi$ denotes the parameters of the model, which includes all the means, covariances, and mixing coefficients for each Gaussian. We used $K=3$ components in our experiments, as done in \citep{zhao2019maximum}. 

To compute the probability that a trajectory is replayed after prioritization, we followed the implementation code which differed from the mathematical derivation in the main paper. Specifically, we first computed a shifted an normalized negative log-likelihood of the trajectories, $q(\tau_i)$:
\begin{equation}
    q(\tau_i) = \frac{-\log p(\tau_i \mid \phi) - c}{\sum_{n=1}^N -\log p(\tau_n  \mid \phi) - c}, \quad \text{where} \quad c = \min_j -\log p(\tau_j  \mid \phi).
\end{equation}
Then we assign the probability of sampling the trajectory, $\bar{q}(\tau_i))$, as the normalized ranking of $q(\tau_i)$:
\begin{equation}
    \bar{q}(\tau_i) = \frac{ \text{rank}(q(\tau_i))}{\sum_{n=1}^N \text{rank}(q(\tau_n))}
\end{equation}
The intuition is that the higher probability trajectory $p(\tau)$ will have a smaller $q(\tau)$, and thus also smaller rank, leading to small probability $\bar{q}(\tau)$ of being sampled. Conversely, lower probability trajectory will be over sampled, leading to an increased entropy of the training distribution. 

\paragraph{Hyperparameter Selection}
\iflong
Many of the parameters above were initially based on what has worked in the past (e.g., many are based on \citet{plappert2018multi} and \citet{baselines}). We found that larger than usual neural networks, inspired by \citet{nair2018overcoming}, accelerate learning slightly, and run efficiently due to our use of GPU. 

To finetune the base agent hyperparameters, we ran two random searches on \env{PointMaze} in order to tune a MEGA agent. These hyperparameters were used for all agents. First, we ran a random search on the \strat{rfaab} ratios. The search space consisted of 324 total configurations:
\begin{itemize}
\itemsep0em 
\item real experience proportion in $\{0, \textbf{1}, 2\}$
\item future relabeling proportion in $\{3, \textbf{4}, 5, 6\}$
\item actual relabeling proportion in $\{1, 2, \textbf{3}\}$
\item achieved relabeling proportion in $\{\textbf{1}, 2, 3\}$
\item behavioral relabeling proportion in $\{0, \textbf{1}, 2\}$
\end{itemize} 
We randomly generated 32 unique configurations from the search space ($\sim 10\%$), ran a single seed of each on \env{PointMaze}, and chose the configuration that converged the fastest---the selected configuration, \strat{rfaab\_1\_4\_3\_1\_1}, is bolded above. We found, however, that using more \strat{future} labels helped in \env{FetchPickAndPlace} and \env{FetchStack2}, for which we instead used \strat{rfaab\_1\_5\_2\_1\_1}.

Our second random search was done in a similar fashion, over various agent hyperparameters:
\begin{itemize}
\itemsep0em 
\item epsilon random exploration in $\{0, \textbf{0.1}, 0.2, 0.3\}$
\item reinforcement learning algorithm in $\{\textrm{\textbf{DDPG}}, \textrm{TD3}\}$ \cite{fujimoto2018addressing}
\item batch size in $\{1000, \textbf{2000}, 4000\}$
\item optimize policy/critic networks every $n$ steps in $\{\textbf{1}, 2, 4, 8\}$
\item go exploration percentage in $\{0., 0.02, 0.05, \textbf{0.1}, 0.2\}$ \cite{ecoffet2019go}
\item action noise in $\{\textbf{0.1}, 0.2\}$
\item warmup period (steps of initial random exploration) in $\{2500, \textbf{5000}, 10000\}$
\end{itemize}
In this case the search space consists of 960 total configurations, from which we generated 96 unique configurations (10\%) and ran a single seed of each on \env{PointMaze}. Rather than choosing the configuration that converged the fastest (since many were very similar), we considered the top 10 best performers and chose the common hyper-parameter values between them. We found that performance is sensitive to the ``optimize every'' parameter, which impacts an agent's stability, and we needed to use less frequent optimization on the more challenging environments. Based on some informal experiments in other environments, we chose to optimize every two steps in \env{Antmaze}, ever four steps in \env{FetchPickAndPlace}, and every ten steps in \env{FetchStack2}.

We additionally ran a third, exhaustive search over the following parameters for the KDE density estimator:
\begin{itemize}
\itemsep0em 
\item bandwidth in $\{0.05, \textbf{0.1}, 0.2, 0.3\}$
\item kernel in $\{\textrm{exponential}, \textrm{\textbf{Gaussian}}\}$
\end{itemize}
This confirmed our initial thought of using the default hyperparameters with normalized goals. 

Finally, to tune our goal discriminator baseline, we ran one seed of each of the following settings for minibatch size and history length $\{(50, 200), (50, 500), (50, 50), \textbf{(100, 200)}, (100, 500), (100, 1000), (20, 200), (20, 500), (20, 50)\}$ (details of these hyperparameters described below).
\else
Omitted in abridged version.
\fi

\iflong
\algrenewcommand\algorithmiccomment[1]{\hfill #1}
\begin{wrapfigure}{R}{0.5\textwidth}
\vspace{-0.3in}
\begin{minipage}{0.5\textwidth}
    \scriptsize
    \setcounter{algorithm}{1}
    \begin{algorithm}[H]
    	\caption{O/MEGA \textsc{Select} functions} \label{alg_mega_select2}
    	\begin{small}
    		\begin{algorithmic}
    		\Function{OMEGA\_Select }{env goal $g_\textrm{ext}$, bias $b$, $*args$}:
    		\State $\alpha \gets 1/\max(b + \DKL{p_{dg}}{p_{ag}}, 1)$
    		\If{$x \sim \mathcal{U}(0, 1) < \alpha$} \Return $g_\textrm{ext}$
    		\Else \ \Return \textsc{MEGA\_Select}($*args$)
    		\EndIf
    		\EndFunction
    		\Statex
    		\Function{MEGA\_Select }{buffer $\mathcal{B}$, num\_candidates $N$}:
    		\State Sample $N$ candidates $\{g_i\}_{i=1}^N \sim \mathcal{B}$
    	    \State Eliminate unachievable candidates (see text)
    		\State \Return $\hat g = \argmin_{g_i} \hat p_{ag}(g_i)$ \Comment{$(*)$}
    		\EndFunction
    		\end{algorithmic}
    	\end{small}
    \end{algorithm}
\end{minipage}
\vspace{-0.3in}
\end{wrapfigure}
\fi

\paragraph{Goal Selection and Baselines} \label{appendix_goaldisc}

\iflong
We reproduce Algorithm \ref{alg_mega_select} to the right, and note that all baselines are simple modifications of line $(*)$ of \textsc{MEGA\_Select}, except the GoalDisc baseline, which does not eliminate unachievable candidates using the cutoff mechanism. All baselines inherit all of the Base Agent features and hyperparameters described above. Then implementation details of each baseline are described below. 
\fi
For MEGA and OMEGA implementations, please refer to Algorithm \ref{alg_mega_select} and the paragraph on Density Modeling above.

\vfill

\iflong
\begin{enumerate}
\item \textbf{Diverse}
\fi

The Diverse baseline scores candidates using $\frac{1}{\hat p_{ag}}$, where $\hat p_{ag}$ is estimated by the density model (KDE, see above), and then samples randomly from the candidates in proportion to their scores. This is similar to using Skew-Fit with $\alpha = -1$ \cite{pong2019skew} or using DISCERN's diverse strategy \cite{warde-farley2018unsupervised}.

\iflong
\item \textbf{Achieved}
\fi

The Achieved baseline samples a random candidate uniformly. This is similar to RIG \cite{nair2018visual} and to DISCERN's naive strategy \cite{warde-farley2018unsupervised}.

\iflong
\item \textbf{GoalDisc}
\fi

This GoalDisc baseline adapts \citet{florensa2018automatic}'s GoalGAN to our setting. To select goals, the Goal Discriminator baseline passes the goal candidates, along with starting states, as input to a trained goal discriminator, which predicts the likelihood that each candidate will be achieved from the starting state. The goals are ranked based on how close the output of the discriminator is to 0.5, choosing the goal with the minimum absolute value distance (the ``most intermediate difficulty'' goal). We do not use the cutoff mechanism based on Q-values in this strategy.

\iflong
To train the discriminator, we start by sampling a batch of 100 of the 200 most recent trajectories (focusing on the most relevant data at the frontier of exploration) (see paragraph on hyperparameters for other considered values). The start state and behavioural goal of each trajectory are the inputs, and the targets are binary values where 1 indicates that the behavioural goal was achieved at some point during the trajectory. The output is continuous valued between 0 and 1, and represents the difficulty of the given start state and behavioural goal combination, in terms of the probability of the agent achieving the goal from the start state. We use the same neural network architecture for the goal discriminator and critic, in terms of layers and neurons; for the discriminator, we apply a sigmoid activation for the final output. We train the discriminator every 250 steps using a binary cross entropy loss.
\fi

\iflong
\item \textbf{MinQ}
\fi

The MinQ strategy uses Q-values to identify the most difficult achievable goals. Candidate goals are ranked based on their Q-values, and the goal with the lowest Q-value is selected. This is similar to DDLUS \cite{hartikainen2020dynamical}.

\iflong
\end{enumerate}
\fi

\iflong
\paragraph{Compute resources}

Each individual seed was run on a node with 1 GPU and $n$ CPUs, where $n$ was between 3 and 12 (as noted above, we made efforts to maintain similar $n$ for each experiment, although unlike the implementation of \citet{plappert2017parameter}, CPU count does not materially impact performance in our implementation, as it only locally shuffles the order of environment and optimization steps), and the GPU was one of \{Nvidia Titan X, Nvidia 1080ti, Nvidia P100, Nvidia T4\}.
\fi

\subsection{Environment Details}

\paragraph{\env{PointMaze}} The \env{PointMaze} environment is identical to that of \citet{trott2019keeping}. The agent finds itself in the bottom left corner of a 10x10 maze with 2-dimensional continuous state and action spaces, and must achieve desired goals sampled from the top right corner within 50 time steps. The state space and goal space are (x, y) coordinates, and the agent moves according to its action, which is a 2-dimensional vector with elements constrained to $(-0.95, 0.95)$. The agent cannot move through walls, which the agent does not see except through experience (i.e., failed actions that press the agent against a wall). Because the agent does not see the walls directly, this is a very difficult exploration environment (indeed, it is the only environment where no seed of any baselines achieves any meaningful success). 

\paragraph{\env{Antmaze}} The \env{Antmaze} environment is identical to that of \citet{trott2019keeping}, which is based on the ant maze environment of \citet{nachum2018data}, which expanded the ant maze used in \citet{florensa2018automatic} by a factor of 4 in each direction (16x the total area). The agent is 3-dimensional ant in a 2-dimensional maze with limits [-4, 20] in both directions. The agent starts on one end of the U-shaped tunnel and must navigate to the desired goal distribution on the other end within 500 timesteps. As compared to \env{PointMaze} the horizon of this environment is significantly longer, but the exploratory behavior required to solve it is considerably simpler, since there are no deadends and only two 90 degree turns are required.

\paragraph{\env{FetchPickAndPlace} (Hard)} The \env{FetchPickAndPlace} environment is based on \env{FetchPickAndPlace-v1} from OpenAI gym \cite{gym2016}, which was first introduced by \citet{andrychowicz2017hindsight}. In this environment the agent is a robotic arm that must lift a block to a desired 3-dimensional target position. Our only change to the environment is to change the desired goal distribution so that all desired goals are between 20cm and 45cm in the air. This is in contrast the easier \env{FetchPickAndPlace-v1}, which has 50\% of all goals on the table, and the other 50\% between 0cm and 45cm in the air. The ``easy'' goal distribution in \env{FetchPickAndPlace-v1} was specifically designed so that a plain HER agent could solve the environment \cite{andrychowicz2017hindsight,plappert2018multi}. To our knowledge, we are the first to solve the hard version without behavioral cloning (as done by \citet{nair2018overcoming}). 

\paragraph{\env{FetchStack2}} The \env{FetchStack2} environment is based on the \env{Fetch} simulation from OpenAI gym \cite{gym2016}, and is intended to replicate the \env{FetchStack2} tasks used by \citet{duan2017one} and \citet{nair2018overcoming}. In fact, our \env{FetchStack2} design is somewhat more difficult than the version used by \citet{nair2018overcoming}. Whereas desired goals in \citet{nair2018overcoming} always place the desired goal above one of the initial block positions (requiring the agent to move only 1 block), our implementation always requires the agent to move both blocks to a new stacked position. The agent has 50 timesteps to move the blocks into position, and success is computed on the last step of the episode (i.e., the blocks must stay stacked). Blocks are initialized in a square of width 6cm about locations (1.3, 0.6) and (1.3, 0.9), and the target position is sampled uniformly in a square of width 30cm in the center of the table. To our knowledge, we are the first to solve this difficult task without demonstrations \cite{duan2017one} or a task curriculum \cite{colas2018curious}.

\iflong
\section{Additional Experiment Details and Experiments}\label{appendix_additional_experiments}

\subsection{Number of seeds and plotting details}

Figures display the average over a set of seeds, with shaded regions representing 1 standard deviation (our plotting script uses the same logic as that of \citet{plappert2018multi} and \citet{baselines}). Figures \ref{fig:main_results} and \ref{fig:main_kde_entropy} display 5 seeds for each setting. Figures \ref{fig_her_horizon}, \ref{fig_density_estimation} and \ref{fig_am_gg} display 3 seeds. Visualizations of achieved goals are all from the same seed. Hyperparameter search was run with 1 seed for each setting, as described above. 

\subsection{FetchPickAndPlace - Increasing Horizon - Details}
\label{appendix_increasinghorizon}

For this experiment, we modified the desired goal distribution of \env{FetchPickAndPlace} to be uniform over the range stated in the legend of Figure \ref{fig_her_horizon} and ran the HER and OMEGA agents (as described above) in the modified environment. 

\subsection{Minimum Density Approximation Versus Learned Conditional for Entropy Gain}
\label{appendix_approxentropy}
In this section we investigate directly learning a conditional model $q(g'|\hat g)$ of the achieved goal $g'$ given the behavioural goal $\hat g$, in order to estimate the entropy gain, and compare it with our minimum density approximation. 

For modeling the conditional $q(g'|\hat g)$, we factorize the conditional using the joint and the behavioural goal models: $q(g'|\hat g) = p(g', \hat{g})/p_{bg}(\hat g)$, both implemented with KDE. For the joint, we sample pairs of achieved and behavioural goals $(g', \hat g)$ from the replay buffer, concatenate them in the input dimension (doubling the dimension), then fit the KDE. For the behavioural goal, we only sample the behaviour goals to train the marginal KDE. Similarly to the KDE for the achieved goal buffer, we also refit the
model on every optimization step using 10,000 normalized samples, sampled uniformly from the respective buffers. We draw joint samples from the replay buffer then compute its conditional probability using the KDE model, instead of a generative approach where we sample potential achieved goals given behaviour goal from a conditional density model. This prevents potential hallucinating of unachievable goals, such as a coordinate outside the \env{PointMaze}.

During behavioural goal selection, we sample $N$ candidates goals $\{g_i\}^N_{i=1} \sim \mathcal{B}$ from achieved goal replay buffer. For each candidate goals, we also sample $K=10$ achieved goals, where one of the goal is the candidate goal itself. Given this set of $\{g'_i, \hat g_j\}_{i=1,...,K,j=1,...,N}$ candidate behaviour goal and achieved goal pairs, we compute the monte carlo estimate of the expected entropy gain for each candidate behaviour $\hat g_j$ using Equation \ref{eq_prop_entropy_gain}.

\begin{figure}
   \centering
  \begin{subfigure}{0.32\textwidth}
    \includegraphics[width=0.95\textwidth]{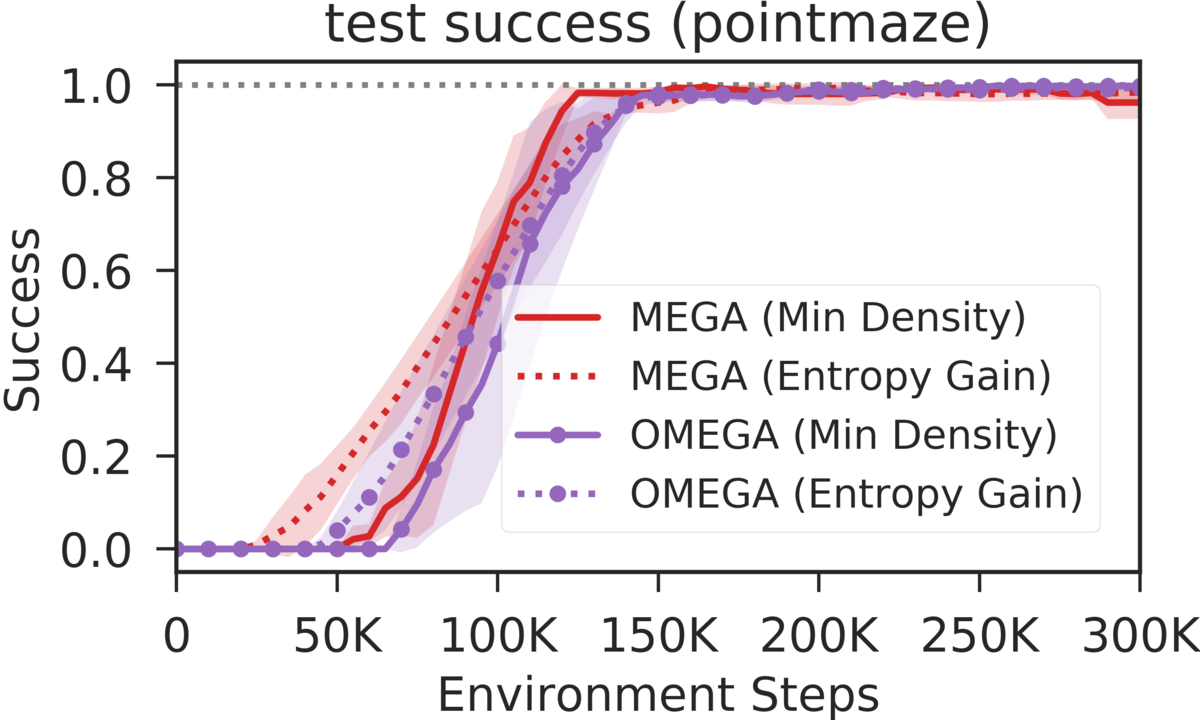}
  \end{subfigure}
  \begin{subfigure}{0.32\textwidth}
    \includegraphics[width=0.95\textwidth]{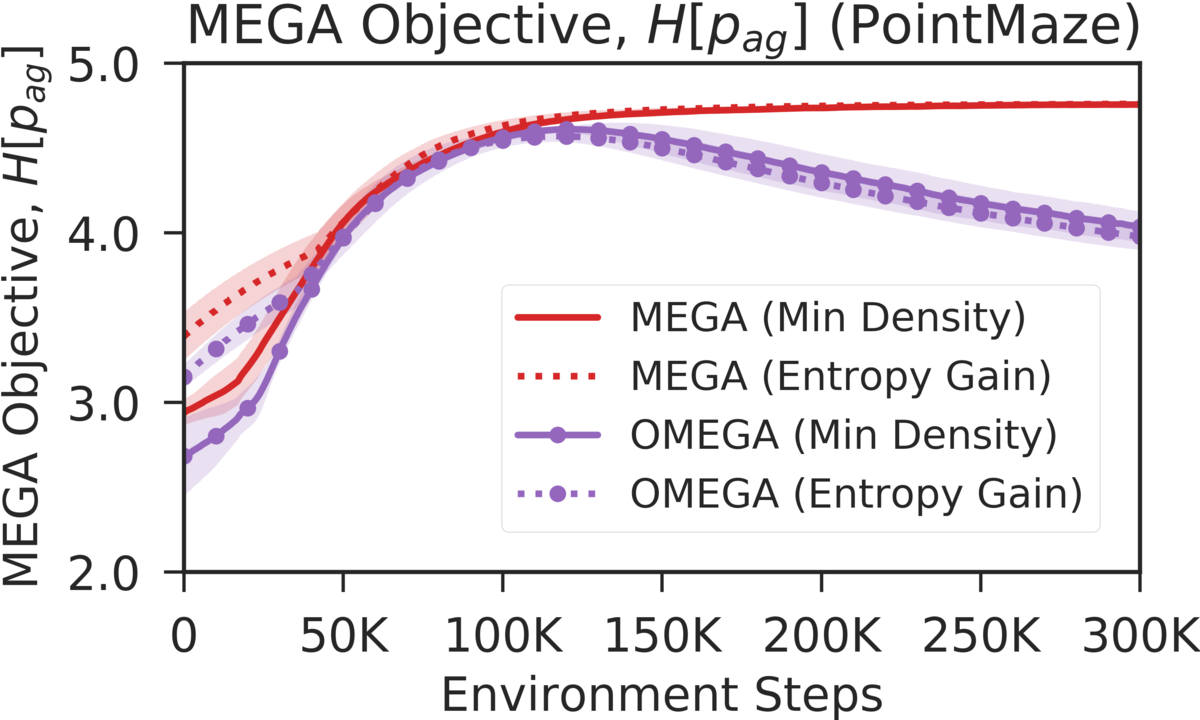}
  \end{subfigure}
  \begin{subfigure}{0.32\textwidth}
    \includegraphics[width=0.95\textwidth]{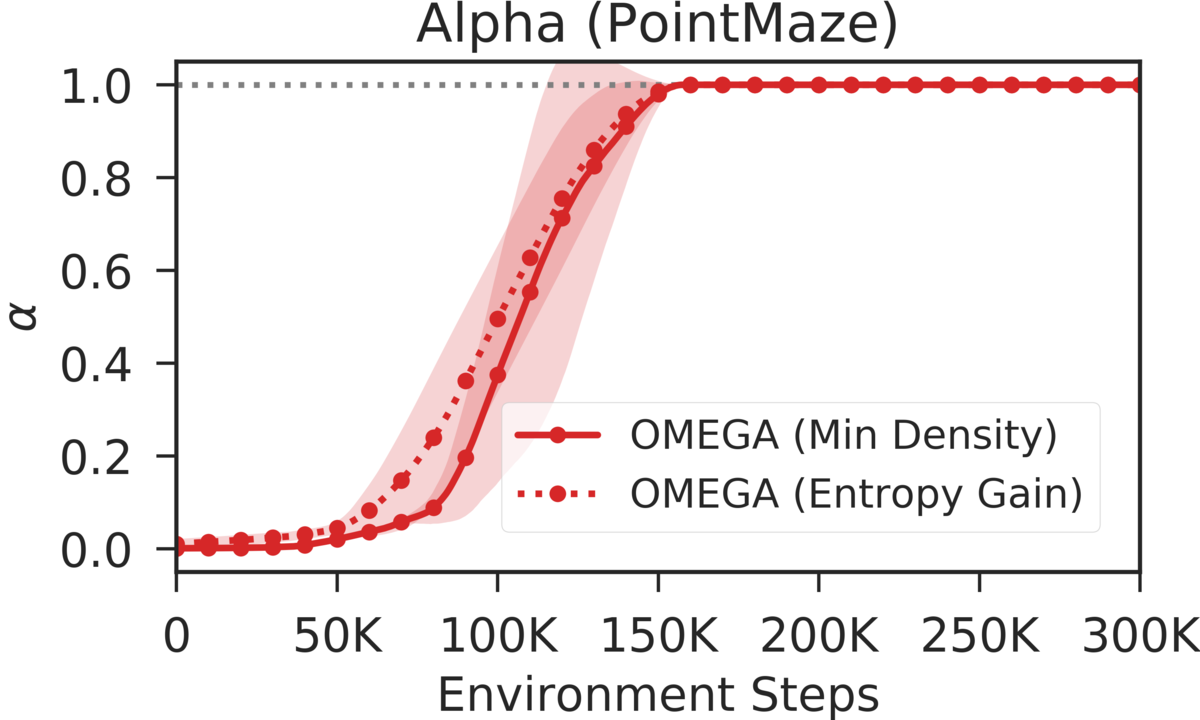}
  \end{subfigure}
  \vspace{-0.1in}
  \caption{\small Comparing with MEGA and OMEGA on (\textbf{Left}) Test Success, (\textbf{Middle}) Achieved Goal buffer KDE estimated entropy, and (\textbf{Right}) $\alpha$ (OMEGA only) when using Minimum Density versus Estimated Entropy Gain on \env{PointMaze}.}\label{fig_pm_mega_eg}
  \vspace{-0.2in}
\end{figure}

We experiment on \env{PointMaze} to compare using minimum density versus learned conditional for estimating entropy gain for choosing behavioural goal. Figure \ref{fig_pm_mega_eg} (left) illustrates that learning the conditional to estimate the entropy gain does lead to slightly faster learning progress (e.g. less environment steps to reach 95\% success rate) than if we use the minimum density sampling. However, because we need to learn and perform inference on the conditional model as well, we find that the our current implementation of the entropy gain approach needs \textit{triple} the wall clock time compared to the simpler minimum density variant of MEGA/OMEGA. 

In theory, choosing the behavioural goals greedily based on the expected entropy gain estimation method should yield higher entropy in the historical achieved goal buffer (MEGA objective), compared to the minimum density approximation. In practice, the approach depends on the the quality of the conditional model $q(g'|\hat g)$ to give estimates of the entropy gain for each behavioural goal. Figure \ref{fig_pm_mega_eg} (middle) shows that initially while the conditional model is being trained on limited data, the entropy of the historical achieved goals (MEGA objective) is actually less than if we simply use the minimum density sampling. However, as the training progresses, the conditional model is able to help estimate more optimal behavioural goals that lead to more entropy gain, overtaking the minimum density sampling approach. 
Finally, Figure \ref{fig_pm_mega_eg} (right) compares the $\alpha$ parameter for OMEGA method, where the entropy gain approach starts to take off later than minimum density, but has a quicker raise to the top, correlating to the test success plot.

Due to the similar test success and MEGA objective performance while being much simpler and faster to run, we perform the rest of the experiments with the minimum density variant of our MEGA/OMEGA method.

\subsection{Ablation of implementation features}\label{apdx_ablation}

As described above, our implementation uses three ``tricks'' that are only indirectly related to our MEGA objective, but that we found to be helpful in certain circumstances. They are: (1) \texttt{rfaab} goal sampling, (2) the use of a Q-value cutoff to eliminate unachievable goal candidates, and (3) an increase in action space exploration upon achieving intrinsic goals during exploration (``goexp''). Figure \ref{fig:ablation} ablates these features on \env{Pointmaze} and \env{FetchStack2}. We observe that \texttt{rfaab} sampling improves performance in both cases relative to HER's \texttt{future\_4} strategy. While the cutoff slightly hurts performance on \env{Pointmaze}, it sometimes prevents the agent from diverging in more complex environment. Finally, we observe that go exploration substantially improves results on \env{Pointmaze}, and sometimes allows the agent to find a solution on \env{FetchStack2} when it otherwise wouldn't have. Note that the \env{FetchStack2} agent uses slightly revised hyperparameters here, which we found to be more stable than the ones used in our main experiments. The differences are: less action l2 regularization (1e-2), more frequent target network updates (every 10 steps), and no gradient value clipping.

\begin{figure}[h]
	\hphantom{xx}\hfill
    \begin{subfigure}[b]{0.31\textwidth}
         \centering
         \includegraphics[width=\textwidth]{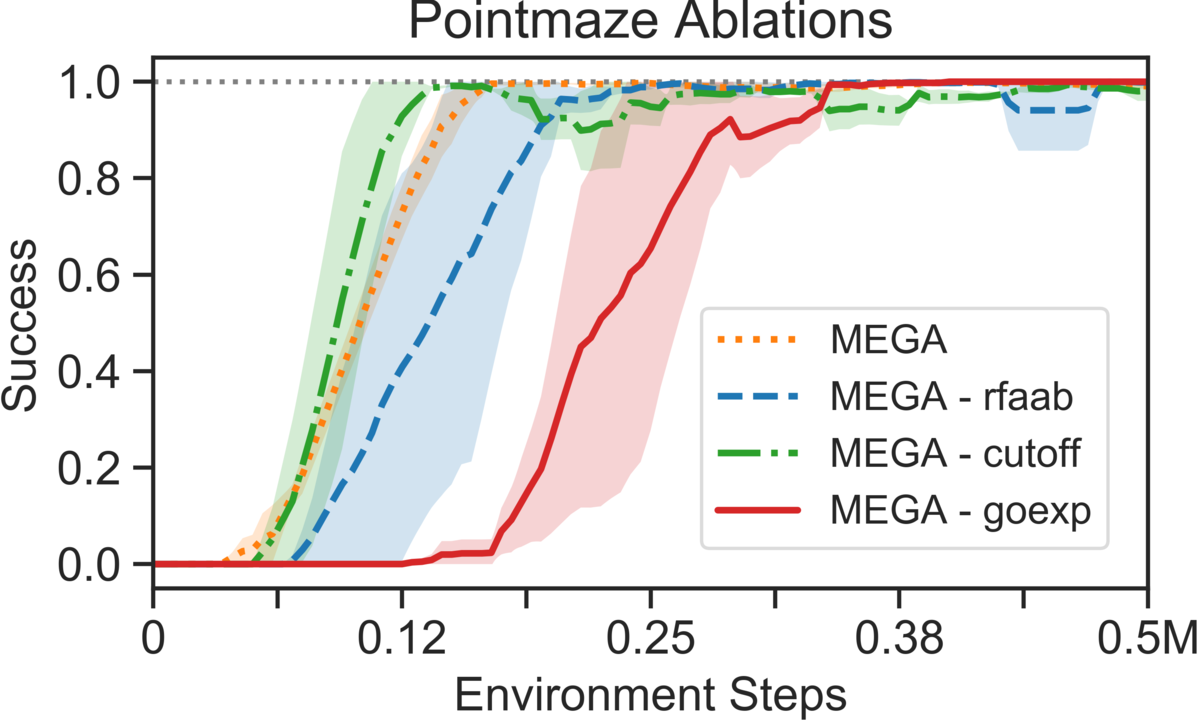}
     \end{subfigure}
     \hspace{0.1in}
     \begin{subfigure}[b]{0.31\textwidth}
         \centering
         \includegraphics[width=\textwidth]{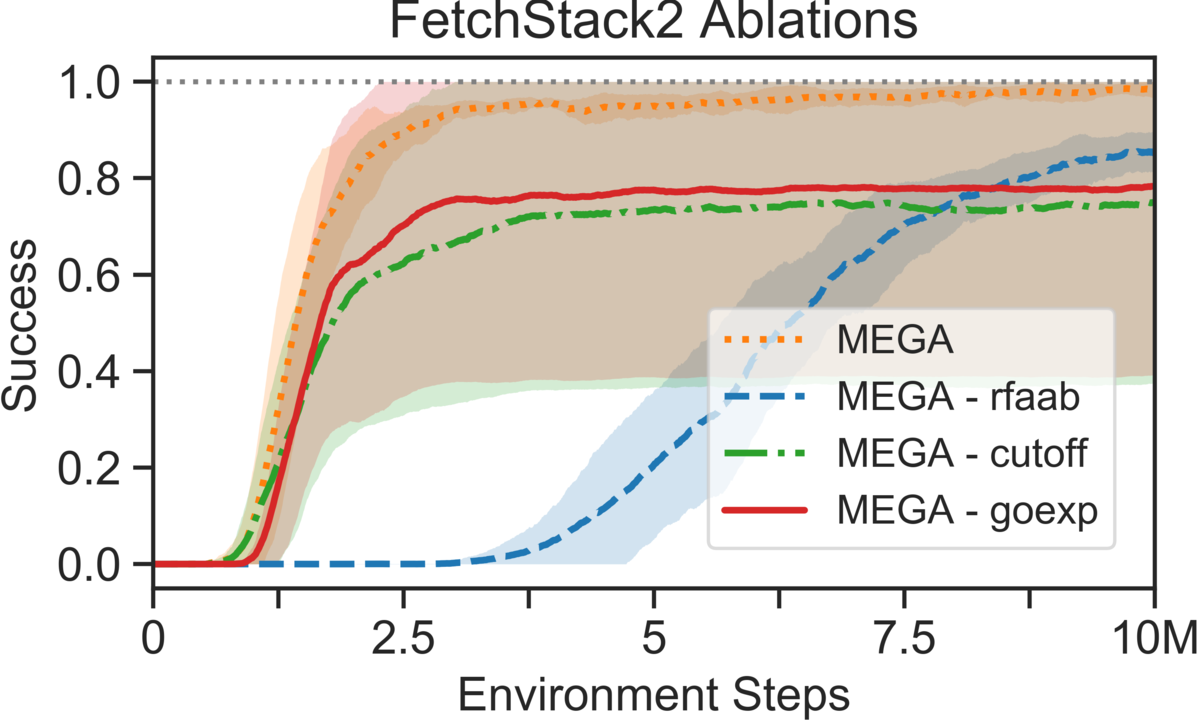}
     \end{subfigure}
     	\hfill
    \vspace{-0.1in}
    \caption{Ablation of certain implementation features in \env{Pointmaze} and \env{FetchStack2}.
    }
    \label{fig:ablation}
\end{figure}

\begin{wrapfigure}{r}{0.33\textwidth}
\vspace{-0.5in}
  \begin{center}
    \includegraphics[width=0.31\textwidth]{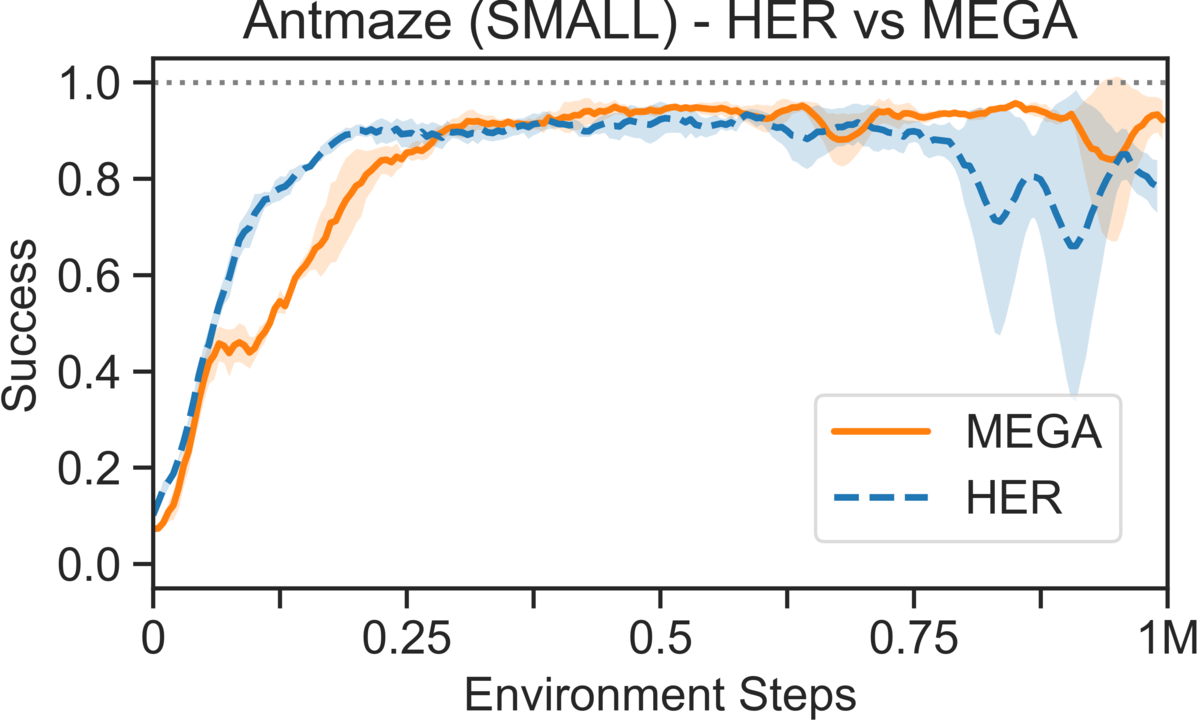}
  \end{center}
  \vspace{-0.2in}
  \caption{\small Small \env{Antmaze} results}\label{fig_am_gg}
\vspace{-0.1in}
\end{wrapfigure}

\subsection{MEGA Tested on Smaller AntMaze}
\label{appendix_smallantmaze}

We also ran HER and MEGA agents on an Antmaze of the same size as used by \citet{florensa2018automatic}. We set the desired goal distribution to be the same as the grid used by \citet{florensa2018automatic} to compute their coverage objective, so that ``success'' in this desired goal space is effectively the same as the objective used by \citet{florensa2018automatic} (note that the HER agent samples directly from this goal distribution during training). The results are shown in Figure \ref{fig_am_gg}. We see that both agents are able to solve this smaller \env{Antmaze} in a fraction of the time required by the TRPO-based agents in \citet{florensa2018automatic} (more than 1000 times faster!). This is more a testament to the power of off-policy learning and goal relabeling than it is to MEGA, since (1) this smaller version is not really a long horizon environment, and (2) our adaptation of \citet{florensa2018automatic}'s GOID (Goals of Intermediate Difficulty) criterion (GoalDisc) solves the larger \env{Antmaze} in approximately the same amount of time as MEGA.  

\subsection{Toy Example: Discrete Entropy Gain} \label{sec:toy1d}

To compare how different strategies optimize the MEGA objective in a controlled environment, we consider an MDP with goal space $g \in \{0,1,...,2n\}$. The initial achieved goal buffer is $\mathcal{B}=\{n\}$. 
At each iteration, the agent chooses behaviour goal $\hat g \in \mathcal{B}$, samples $g' \sim q(g' \given \hat g)$ from the induced conditional, and adds $g'$ to $\mathcal{B}$, where $q(g'|\hat{g})$ 
is defined as:
\begin{align*}\small
    q(g'|\hat{g}) = 
    \begin{cases}
        0.4 & \text{if } g' = \hat{g} \\
        0.2 & \text{if } g' = \hat{g} \pm 1 \\
        0.1 & \text{if } g' = \hat{g} \pm 2, \\
        0   & \text{otherwise}.
    \end{cases}
\end{align*}
At the boundaries, we truncate the $q(g' \given \hat g)$ and normalize. 

We consider four policies for choosing $\hat{g}$ at each iteration: 
\begin{enumerate}
    \item $\pi_\textrm{Achieved}(g) = p_{ag}(g)$, which samples uniformly from the buffer, as used, approximately, by RIG \cite{nair2018visual}
    \item $\pi_\textrm{Diverse}(g) \propto p_{ag}(g) \cdot p_{ag}(g)^{-1}$, which is equivalent to sampling uniformly on the support set of the buffer, as approximated by DISCERN's \cite{warde-farley2018unsupervised} ``diverse'' strategy and Skew-Fit \cite{pong2019skew} (Skew-Fit parameterizes the exponent with $\alpha \in [-1,0)$)
    \item $\pi_\textrm{MEGA}(g) \propto \mathds{1}[g = \arg \min p_{ag}(g)]$ (ours), which samples from the goal(s) with the minimum density or mass
    \item $\pi_\textrm{EG}(g) = \mathds{1}[g = \arg \min L(\hat g)]$, which is an oracle that chooses $\hat{g}$ to maximize the next step entropy gain (\ref{eq_prop_entropy_gain})
\end{enumerate}

Figure \ref{fig:toy_ent_gain} plots empirical entropy over iterations, where $n = 50$, averaged over 50 trials. Our minimum density heuristic is almost as fast as the oracle $\pi_\textrm{EG}$, and converges to the max entropy distribution much faster than $\pi_\textrm{Achieved}$ and $\pi_\textrm{Diverse}$. 

\begin{figure}[h!]
    \centering
    \vspace{-0.1in}
    \begin{subfigure}{0.45\textwidth}
        \includegraphics[width=\textwidth]{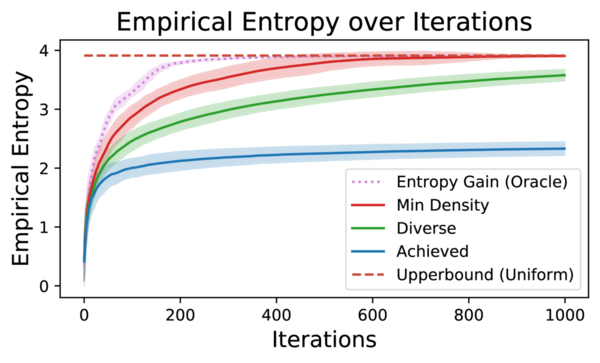}
        \label{fig:toy_ent_gain_entropy}
    \end{subfigure}
    \begin{subfigure}{0.45\textwidth}
        \includegraphics[width=\textwidth]{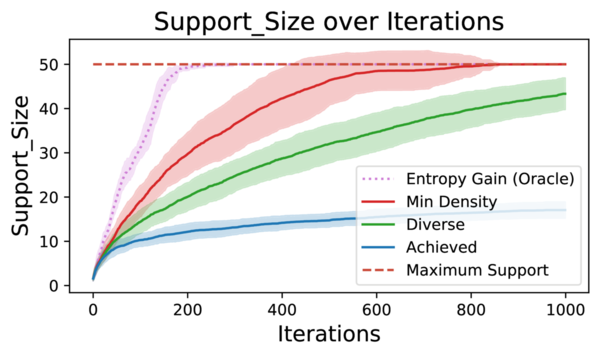}
        \label{fig:toy_ent_gain_support}
    \end{subfigure}
    \vspace{-0.1in}
    \caption{(\textbf{Left}) Empirical entropy of the buffer over time for different goal selection strategies in toy example of Section \ref{sec:toy1d}. (\textbf{Right}) Support size of the buffer data over time steps with different behaviour goal selection strategies in the discrete goal toy example in Section \ref{sec:toy1d}. The curves plots the mean averaged over 50 runs and shades one standard deviation in each direction.}
    \label{fig:toy_ent_gain}
    \vspace{-0.1in}
\end{figure}

We also visualize the empirical probability distribution of the achieved goal buffer over iterations in Figure \ref{fig:toy_probs}, when using the oracle entropy gain sampling $\pi_{EG}(g)$ (top) which has access to ground truth conditional $q(g' \given \hat g)$, versus using the minimum density $\pi_{MEGA}(g)$ (bottom). With the oracle $\pi_{EG}(g)$, the support size (number of non-white rows at a given vertical slice) grows consistently at each iteration. The minimum density $\pi_{MEGA}(g)$ approach does spend time sampling goals in the ``interior" section to ``even out" the distribution (e.g. around iteration 55 to 100) without increasing the support size, before discovering new unseen goals that increase the support size.

\begin{figure}[h!]
    \centering
    \includegraphics[width=0.95\textwidth]{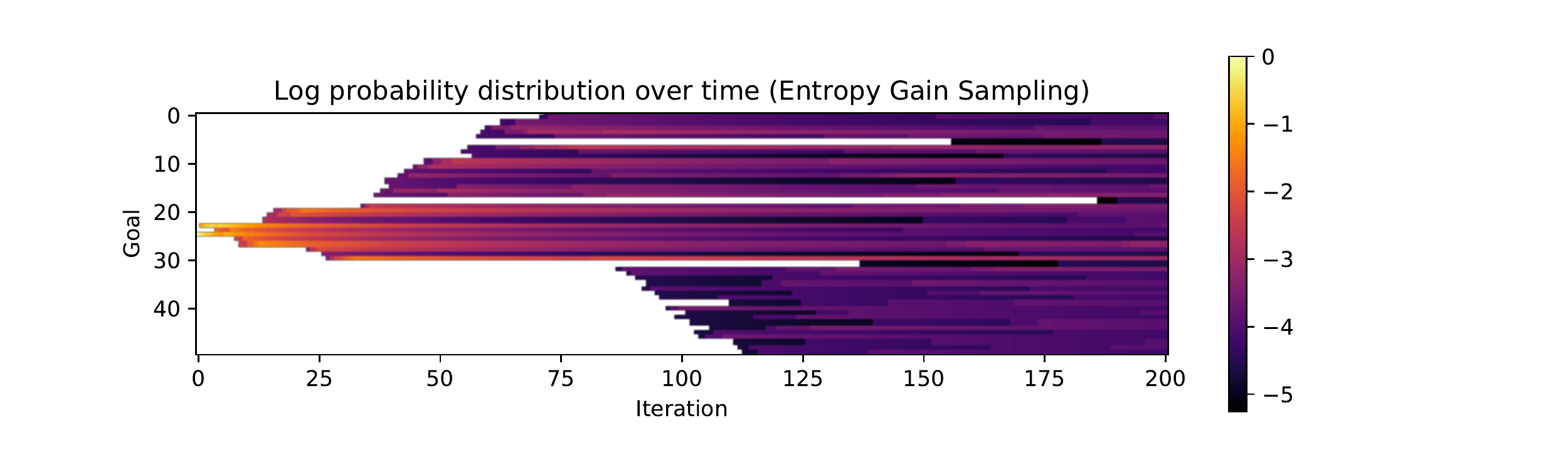}
    \includegraphics[width=0.95\textwidth]{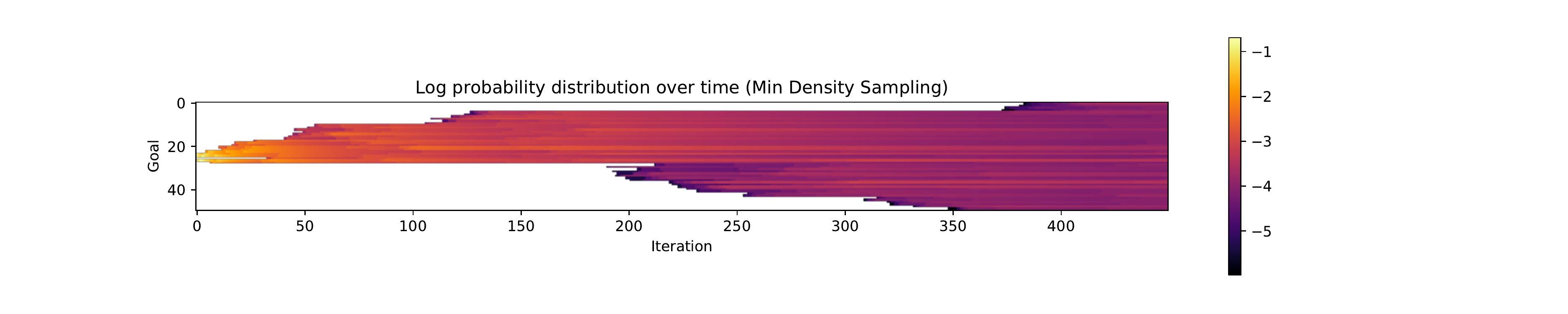}
    \caption{Log probability (denoted by the color) for each achieved goal ($n=50$ possible goals represented in the y-axis) versus iterations for (top) entropy gain sampling, (bottom) minimum density sampling. Note the difference in the scale of the iteration axis between the two plots. Goals with zero probability (not in support) are shaded white. Best viewed in colour.}
    \label{fig:toy_probs}
\end{figure}

\else

\section*{Appendices D, E omitted in abridged.}
\fi

\iflong%
\else
\end{multicols}
\fi

\end{document}

\end{document}